\documentclass[journal]{IEEEtran}

\usepackage{amsmath,amsfonts}
\usepackage{algorithmic}
\usepackage{algorithm}
\usepackage{array}
\usepackage[caption=false,font=normalsize]{subfig}
\usepackage{textcomp}
\usepackage{stfloats}
\usepackage{url}
\usepackage{hyperref}
\usepackage{verbatim}
\usepackage{graphicx}
\usepackage{cite}
\usepackage{colortbl}
\usepackage{booktabs}
\usepackage{multirow}
\usepackage{sistyle}
\SIthousandsep{,}

\usepackage{xcolor}

\hypersetup{
    colorlinks=true,
    citebordercolor=green,
    linkbordercolor=red,
    urlbordercolor=cyan,
}

\def\red#1{\textcolor{red}{#1}}

\usepackage{float}

\usepackage[switch]{lineno}

\usepackage{orcidlink}

\usepackage[normalem]{ulem}
\usepackage{color}

\newcommand{\Remove}[1]{\if0{#1}\fi}  
\newcommand{\Redtext}[1]{\textcolor{red}{{#1}}}  

\DeclareRobustCommand{\redsout}{\bgroup\markoverwith{\Redtext{\rule[.5ex]{2pt}{0.6pt}}}\ULon}  

\newcommand{\RobustSout}[1]{%
    \relax
    \ifmmode
        \hbox{\redsout{$#1$}}
    \else%
        \redsout{#1}%
    \fi%
}

\newcounter{OutputVersion}
\newcommand{\Erase}[1]{%
    \ifnum\value{OutputVersion}=0%
        {#1}%
    \else%
        \ifnum\value{OutputVersion}=2%
            \RobustSout{#1}%
        \else%
            \Remove{#1}%
        \fi%
    \fi%
}

\newcommand{\Add}[1]{%
    \ifnum\value{OutputVersion}=0%
        \Remove{#1}%
    \else%
        \ifnum\value{OutputVersion}=1%
            {#1}%
        \else%
            \Redtext{#1}%
        \fi%
    \fi%
}

\newcommand{\Replace}[2]{%
    \ifnum\value{OutputVersion}=2%
        {\Erase{#1} \Add{#2}}
    \else%
        {\Erase{#1}\Add{#2}}%
    \fi%
}
\begin{document}


\title{Joint Learning of Blind Super-Resolution and\\ Crack Segmentation for Realistic Degraded Images}

\author{Yuki~Kondo$^{\orcidlink{0000-0002-5263-8722}}$~and~Norimichi~Ukita$^{\orcidlink{0000-0002-0240-1065}}$,~\IEEEmembership{Member,~IEEE}
\IEEEcompsocitemizethanks{\IEEEcompsocthanksitem Y. Kondo and N. Ukita with Intelligent Information Media Lab, Toyota Technological Institute (TTI), Nagoya, Japan, 468-8511.
(E-mail: yuki\_kondo@toyota-ti.ac.jp; ukita@toyota-ti.ac.jp)
}

\thanks{This code will be released in \url{https://github.com/Yuki-11/CSBSR} after publication.}
\thanks{Manuscript received: September 7, 2023; Revised: December 23, 2023; Accepted: February 18, 2024.}
}

\markboth{Journal of \LaTeX\ Class Files,~Vol.~14, No.~8, August~2021}%
{Shell \MakeLowercase{\textit{et al.}}: A Sample Article Using IEEEtran.cls for IEEE Journals}

\IEEEpubid{\begin{minipage}{\textwidth}\ \\[12pt] \centering
0000--0000/00\$00.00~\copyright~2024 IEEE.  Personal use of this material is permitted.  Permission from IEEE must be obtained for all other uses, in any current or future media, including reprinting/republishing this material for advertising or promotional purposes, creating new collective works, for resale or redistribution to servers or lists, or reuse of any copyrighted component of this work in other works.
\end{minipage}}

\maketitle

\begin{abstract}
This paper proposes crack segmentation augmented by super resolution (SR) with deep neural networks.
In the proposed method, a SR network is jointly trained with a binary segmentation network in an end-to-end manner.
This joint learning allows the SR network to be optimized for improving segmentation results.
For realistic scenarios, the SR network is extended from non-blind to blind for processing a low-resolution image degraded by unknown blurs.
The joint network is improved by our proposed two extra paths that further encourage the mutual optimization between SR and segmentation.
Comparative experiments with State of The Art (SoTA) segmentation methods demonstrate the superiority of our joint learning, and various ablation studies prove the effects of our contributions.
\end{abstract}

\begin{IEEEkeywords}
Crack segmentation, Image processing, Blind super-resolution, Joint learning, Multi-task learning
\end{IEEEkeywords}

\IEEEpubidadjcol
\section{Introduction}
\label{section:introduction}

\IEEEPARstart{W}{hile} many constructions and infrastructures such as buildings, pavements, bridges, and tunnels are dilapidated in the world, it is difficult to always manually inspect all of them.
Instead of the manual inspection, automatic inspection is one of the prospective solutions for efficiently diagnosing these constructions.
While such inspection can be achieved by several types of sensors such as the Falling Weight Deflectometer, the Pavement Density Profiler, and the Ground Penetrating Radar, this paper focuses on crack segmentation on images captured by generic cameras for visual inspection.

\begin{figure*}[!t]
\begin{footnotesize}
\begin{center}
\includegraphics[width=\textwidth]{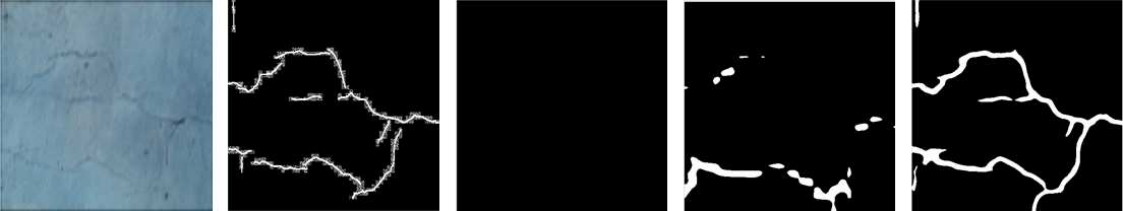}
~\hspace*{1mm}
(a) Input LR \hspace*{19mm} (b) HR GT  \hspace*{18mm}
(c) Independent~\cite{bib:SrcNet} \hspace*{14mm}
(d) Multi-task~\cite{DBLP:conf/cvpr/WangLZTS20} \hspace*{14mm}
(e) Ours (CSBSR)
\end{center}
\caption{Crack segmentation challenges for synthetically-degraded images given by low resolution and anisotropic Gaussian blur (same experimental conditions as in section~\ref{subsection:exp_detail}). From an input degradated LR image (a), High-Resolution (HR) segmentation results (c), (d), and (e) are acquired. (c) Independent and (d) Multi-task show the results on images enlarged by non-blind SR ``trained independently of segmentation'' and ``trained with segmentation in a multi-task learning manner,'' respectively.
(b) is the manually-annotated ground-truth (GT) HR segmentation image. In (c), the independently optimized non-blind SR model does not allow sufficient image enhancement to make the segmentation model easy to infer, and the segmentation model does not detect cracks. (d) can detect some cracks, but there are undetected cracks. Our method (e) succeeds in detecting cracks in the most detail.}
\label{fig:top}
\end{footnotesize}
\end{figure*}

Crack segmentation~\cite{bib:crack_survey1} is defined to be binary semantic segmentation in the field of computer vision.
While the number of classes in crack segmentation (i.e., two classes) is much fewer than 
the recent generic multiclass segmentation~\cite{DBLP:journals/pami/ShelhamerLD17, DBLP:journals/pami/HeGDG20, DBLP:conf/cvpr/KirillovHGRD19}, real-world crack segmentation is not
an easy problem even with recent powerful deep neural networks.
This is because of the following reasons:
\begin{enumerate}
\renewcommand{\labelenumi}{$\langle$\textit{\Alph{enumi}}$\rangle$.}
\item {\bf High class-imbalance:} The number of crack pixels is much less than the number of non-crack pixels (i.e., background pixels), as shown in Fig.~\ref{fig:top} (b).
In such a problem, all pixels tend to be classified to background.
\item {\bf Fine cracks:} Cracks can be hairline,
which are difficult to be segmented, as shown in Fig.~\ref{fig:top} (b).
\item {\bf Low-Resolution (LR):} For inspection of various structures such as tunnels~\cite{DBLP:conf/mva/StentGSSC13} and pavements~\cite{DBLP:journals/tits/FeiWZCLLYL20}, an inspection camera captures cracks in LR (as shown in Fig.~\ref{fig:top} (a)) because it cannot get close to the structures for safety reasons.
\item {\bf Cracks in blurred images:} Since inspection images are usually captured from moving vehicles such as cars and drones for efficient inspection, those images can be blurred, as shown in Fig.~\ref{fig:top} (a).
\end{enumerate}

While even each of the aforementioned problems is not an easy problem,
crack segmentation is more challenging due to the combination of all of these problems, even with SoTA methods, as shown in Fig.~\ref{fig:top} (c) and (d).
To cope with these problems, this paper proposes a unified framework consisting of the following novel contributions (Table~\ref{table:contributions}):

\begin{enumerate}
\item {\bf Crack Segmentation with Blind Super-Resolution (CSBSR):} 
As with Crack Segmentation with Super Resolution (CSSR) proposed in our earlier conference paper~\cite{CSSR2021}, CSBSR proposed in this paper connects ``a network for SR accepting an input LR image'' in series to ``a segmentation network'' for end-to-end joint learning.
We extend CSSR to CSBSR with blind SR to handle realistically-blurred images.
Our joint learning of blind SR and segmentation allows us to optimize SR for improving segmentation (Fig.~\ref{fig:top} (e)) more than similar methods~\cite{bib:SrcNet,DBLP:conf/cvpr/WangLZTS20} using both  non-blind SR and segmentation (Fig.~\ref{fig:top} (c) and (d)).
\item {\bf Boundary Combo (BC) loss for segmentation:} In addition to super-resolving tiny cracks as mentioned above, fine boundaries are locally evaluated with global constraints in the whole image for detecting fine cracks robustly to the class-imbalance problem.
\item {\bf Segmentation-aware SR-loss weights:}
While CSSR and CSBSR use BC loss to train not only the segmentation network but also the SR network in and end-to-end manner, the SR network is less optimized due to gradient vanishing through the segmentation network.
To train the SR network more for segmentation, BC loss directly weights a loss for SR.
For further improvement, the SR loss is also weighted by additional weights based on fine-crack and hard-negative pixels.
\item {\bf Blur skip for blur-reflected task learning:} 
Since an SR image is imperfect, blur effects remaining in the SR image give a negative impact on segmentation.
For segmentation more robustly to the blur effects, the blur estimated in SR is provided to the segmentation network via a skip connection.
\end{enumerate}
\vspace{-1em}
\IEEEpubidadjcol


\begin{table*}[!t]
  \centering
  \small
  \caption{Problems $\langle A \rangle$, $\langle B \rangle$, $\langle C \rangle$, and $\langle D \rangle$ and their solutions 1, 2, 3, and 4.
  If a solution is for a given problem, it is represented by \textcolor{red}{$\surd$} in the table.}
  \label{table:contributions}
    \begin{tabular}{lcccc}
    \toprule
    {} & $\langle A \rangle$ Class imbalance & $\langle B \rangle$ Fine cracks & $\langle C \rangle$ LR cracks & $\langle D \rangle$ Blur \\
    \midrule
    1. CSBSR & & & \textcolor{red}{$\surd$} & \textcolor{red}{$\surd$} \\
    \rowcolor[rgb]{.851, .851, .851}
    2. BC loss & \textcolor{red}{$\surd$} & \textcolor{red}{$\surd$} & & {} \\
    3. Segmentation-aware SR-loss weights & & \textcolor{red}{$\surd$} & \textcolor{red}{$\surd$} & {} \\
    \rowcolor[rgb]{.851, .851, .851}
    4. Blur-reflected task learning & & & & \textcolor{red}{$\surd$} \\
    \bottomrule
    \end{tabular}
  \label{tab:addlabel}
\end{table*}

\section{Related Work}
\label{section:related}

\subsection{Image Segmentation}
\label{subsection:segmentation}

Image segmentation techniques~\cite{DBLP:journals/corr/abs-2001-05566} are briefly divided into three categories, namely semantic segmentation~\cite{DBLP:journals/pami/ShelhamerLD17}, instance segmentation~\cite{DBLP:journals/pami/HeGDG20}, and panoptic segmentation~\cite{DBLP:conf/cvpr/KirillovHGRD19}.
Crack segmentation is categorized into semantic segmentation because it classifies all pixels into crack and background pixels with no instance.
That is, these crack pixels are not divided into crack instances.

\noindent
{\bf Class-imbalance Segmentation:}
As well as in various computer vision problems, in image segmentation, class imbalance is a critical problem.
Many approaches for class imbalance are applicable to class-imbalance segmentation tasks.
For example, weighted loss such as the Weighted Cross Entropy (WCE) loss~\cite{bib:WCE_crack} and the focal loss~\cite{DBLP:conf/iccv/LinGGHD17} for segmentation~\cite{DBLP:journals/tmi/LiKG21,DBLP:journals/ijon/HossainBP21}, re-sampling for segmentation~\cite{DBLP:conf/cvpr/BuloNK17}, and
hard mining for segmentation~\cite{DBLP:journals/cvm/GongZZYX22}.

Among all segmentation tasks, medical image segmentation has to cope with highly-imbalanced classes (e.g., tiny tumors and background).
Such difficult medical image segmentation is tackled by a variety of loss functions such as the Dice loss~\cite{bib:Dice},
the Generalized Dice loss~\cite{DBLP:conf/miccai/SudreLVOC17}, 
the Combo loss~\cite{bib:taghanaki2019combo}, the Hausdorff loss~\cite{DBLP:journals/tmi/KarimiS20},
and the Boundary loss~\cite{DBLP:journals/mia/KervadecBDGDA21}.

\noindent
{\bf Crack Segmentation:}
Since the class-imbalance issue is important also for crack segmentation as presented as Problem $\langle A \rangle$ in Table~\ref{tab:addlabel}, the aforementioned schemes proposed against class imbalance are useful for crack segmentation.
For example, in order to balance the number of samples between classes, K. Zhang et al.~\cite{bib:Zhang2020} oversamples crack images.
The Dice, Combo, and WCE losses are employed for crack segmentation 
in~\cite{bib:rezaie2020}, in~\cite{bib:chen2021}, and in~\cite{bib:liu2019deepcrack, bib:liu2021crackformer}, respectively.

In addition to the class-imbalance issue, the fine boundaries of cracks are not easy to be extracted
and make crack segmentation difficult, as presented as Problem $\langle B \rangle$ in Table~\ref{tab:addlabel}.
For such difficult fine crack segmentation, the aforementioned schemes proposed against class imbalance (e.g., weighted loss, re-sampling, class-imbalance-oriented loss) are also useful.
Previous methods for such fine cracks are divided into the following two approaches, namely boundary-based and coarse-to-fine weighting.

In the boundary-based approach, the distance between the boundaries of ground-truth (GT) and predicted cracks is minimized.
In~\cite{DBLP:journals/tmi/KarimiS20}, the Hausdorff distance is evaluated by using the distance transform.
While its computational cost for the exact solution is high, the sum of L2 distances is approximated by the sum of regional integrals for efficiency in the Boundary loss~\cite{DBLP:journals/mia/KervadecBDGDA21}.


Various coarse-to-fine weighting approaches such as~\cite{DBLP:journals/pami/ChenPKMY18} employ pyramid and U-net~\cite{bib:Ronneberger2015unet} like networks for weighting a fine but unreliable representation by more reliable results in a coarse representation.
The effectiveness of this approach is validated also in crack segmentation~\cite{bib:Yang2020, bib:liu2019deepcrack,bib:liu2021crackformer}.

While the effectiveness of the both approaches is validated, the coarse-to-fine weighting approach is applicable only to pyramid and U-net like architectures.
On the other hand, the boundary-based approach can be employed with any other loss functions in any network architectures in general.

\subsection{Super Resolution (SR)}
\label{subsection:SR}

\noindent
{\bf Non-blind SR:} 
SR reconstructs a High-Resolution (HR) image $\boldsymbol{I}^{H} \in \mathbb{R}^{3\times h\times w}$, from its LR image $\boldsymbol{I}^{L} \in \mathbb{R}^{3\times \frac{h}{s}\times\frac{w}{s}}$, where $w \in \mathbb{R}, h \in \mathbb{R},$ and $s \in \mathbb{R}$ are width, height and a scaling factor of the image, respectively.
The image degradation process from HR to LR is modeled as follows\footnote{In principle, in this paper, tensors, including matrices, are denoted by bold italic uppercase letters, vectors by bold italic lowercase letters, and scalars by thin italic lowercase letters. However, in some cases, numbers of pixels and specific samples, sets and functions, such as loss functions and network models, are written in narrow uppercase capital letters.}:
\begin{equation}
\boldsymbol{I}^{L} = (\boldsymbol{I}^{H} \ast \boldsymbol{K})\downarrow_{s},
\label{eq:down_model}
\end{equation}
where $\boldsymbol{K} \in \mathbb{R}^{k\times k}$, $k \in \mathbb{R}$, $\ast$, and $\downarrow_{s}$ denote a blur kernel, a kernel size, a channel-wise convolution operator, and a downsampling process with inverse of magnification factor $s \in \mathbb{R}$, respectively.
By downscaling HR training images to their LR images by a known downsampling process,
such as bicubic interpolation, 
we can have a set of $\boldsymbol{I}^{H}$ and $\boldsymbol{I}^{L}$ for training a non-blind SR model.

Such non-blind SR is developed with various aspects~\cite{bib:wang2021SR-survey} such as arbitrary image degradations~\cite{DBLP:conf/cvpr/ZhangGT20}, recurrent/iterative networks~\cite{DBLP:conf/cvpr/LeePLMKLKHK20}, and reference-based SR~\cite{DBLP:conf/cvpr/LuLTLJ21}.
However, since the image degradation process is assumed to be known in all of these non-blind SR methods, their performance is decreased in real-world images with
arbitrary unknown degradations.

\noindent
{\bf Blind SR:} To apply SR to arbitrarily-degraded images, blur kernel $\boldsymbol{K}$ is employed in blind SR.
Even without modeling $\boldsymbol{K}$ in a SR network, blind SR can be done by blurring training images~\cite{bib:MZSR, bib:BSRGAN2021} or by deblurring input images~\cite{bib:hussein2020correction} by $\boldsymbol{K}$.
In the kernel conditioning approach~\cite{DBLP:conf/cvpr/GuLZD19, DBLP:conf/cvpr/WangWDX0AG21}, a blur representation estimated from an input LR image is fed into a SR network for conditioning the SR process by the blur.
While this kernel conditioning employs low-dimensional blur representations for efficiency
and stability in general, the original blur kernel, $\boldsymbol{K}$, is modeled within a SR network for further accuracy in~\cite{DBLP:conf/cvpr/GuoCWCCDXT20,DBLP:conf/cvpr/KimSK21,
bib:kbpn}.

Since the blur kernel is more informative than its low-dimensional representation, the blur kernel can be useful for additional tasks using a SR image.
As such an additional task, image segmentation is done in our work.

\subsection{Joint Learning of SR and Other Tasks}
\label{subsection:joint}

\begin{figure}[t]
\begin{center}
\vspace{-1.5em}
\subfloat[{\bf {\em Independent}} learning with {\bf {\em non-blind}} SR~\cite{bib:SrcNet}]{\includegraphics[width=\linewidth]{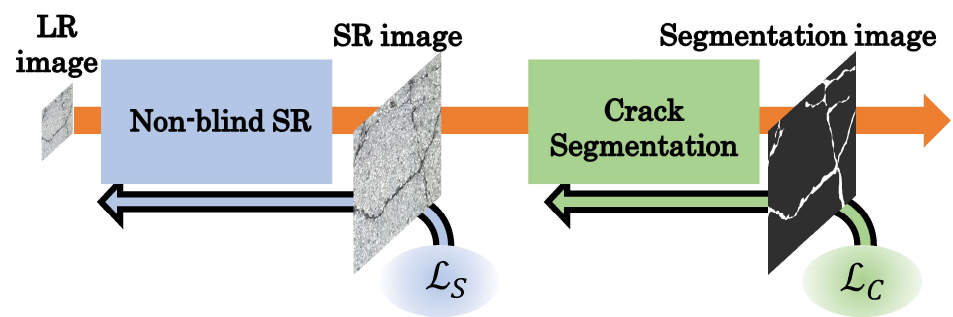}%
\label{subfig:indipendent_learning}}
\vspace{-0.5em}

\subfloat[{\bf {\em Multi-task}} learning with {\bf {\em non-blind}} SR~\cite{DBLP:conf/cvpr/WangLZTS20}]{\includegraphics[width=0.8\linewidth]{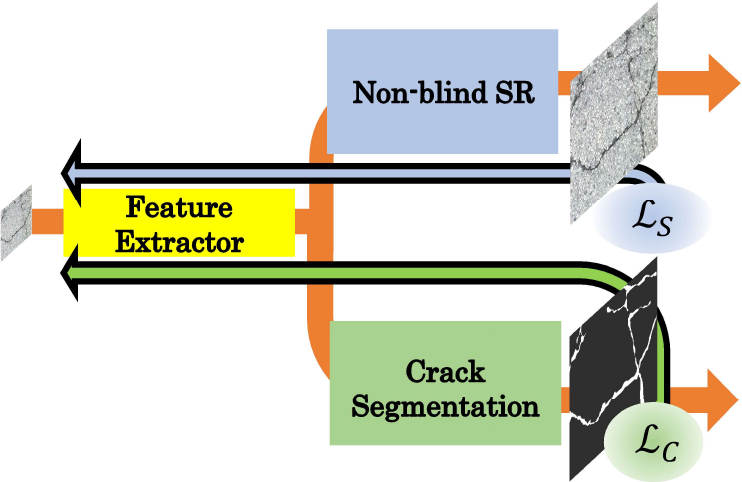}%
\label{subfig:multi-task_learning}}
\vspace{-0.3em}

\subfloat[{\bf {\em Joint}} learning with {\bf {\em blind}} SR and extra paths (Ours)]{\includegraphics[width=\linewidth]{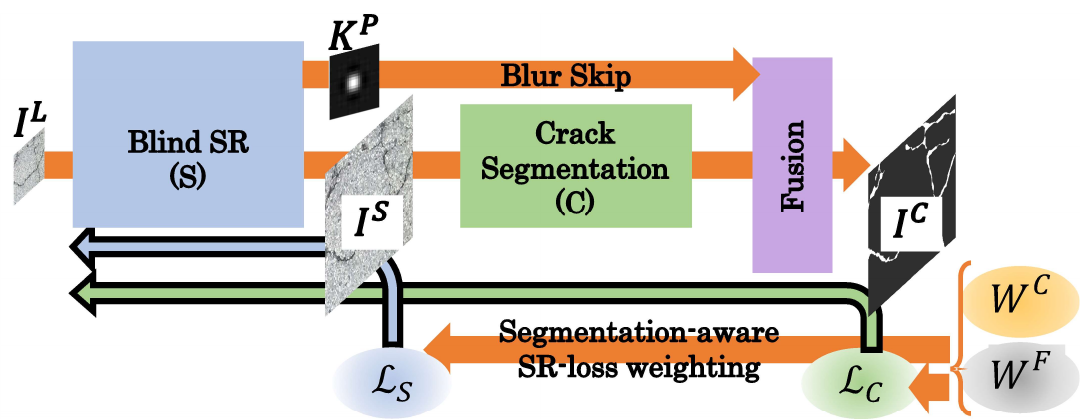}%
\label{subfig:our_method_compe}}
\end{center}
\caption{Combinations of SR and segmentation. (a) Independent learning with non-blind SR~\cite{bib:SrcNet}. (b) Multi-task learning with non-blind SR~\cite{bib:SrcNet}. (c) Our joint learning with blind SR and extra paths called CSBSR.
While orange arrows indicate data flows,
arrows leading out of the loss functions (i.e., $\mathcal{L}_{S}$ and $\mathcal{L}_{C}$) indicate the back-propagation paths for training.
Blue and green arrows indicate the back-propagations given by the SR and segmentation tasks, respectively.
Each ellipse indicates a loss or weights given to a certain loss.
Our CSBSR is illustrated more in detail in Fig.~\ref{fig:network}.
}
\vspace{-1.5em}
\label{fig:joint_learning}
\end{figure}

With upscaled SR images, a variety of applications can be realized.
For example, distant-object
detection~\cite{DBLP:conf/cvpr/HaoLQYLH17}, segmentation~\cite{DBLP:journals/access/GuoWSYCZSXXSS19}, and wide-angle image analysis\cite{DBLP:journals/tits/ChoiCCS18
}. 
As with these examples, crack segmentation can be also supported by SR for detecting LR cracks presented in Problems $\langle C \rangle$ and $\langle D \rangle$ in Table~\ref{tab:addlabel}.

While these methods have models for SR and another task (e.g., segmentation) separately (Fig.~\ref{fig:joint_learning} (a)), these tasks can be jointly trained in a single model for supporting the additional task more explicitly.
Such joint end-to-end learning is also applicable in a variety of tasks such as classification~\cite{DBLP:conf/iccv/SinghN0V19} and detection (e.g., pedestrians~\cite{DBLP:journals/tnn/ZhangBDXG21}, vehicles~\cite{DBLP:conf/ipas2/AkitaHU20}, and generic objects~\cite{bib:TDSR2021}).

Image segmentation can be also improved by combining with SR.
As shown in Fig.~\ref{fig:joint_learning} (b), Dual Super-Resolution Learning (DSRL)~\cite{DBLP:conf/cvpr/WangLZTS20} applies multi-task learning to the non-blind SR and segmentation tasks
so that a single feature extractor is shared by the parallel SR and segmentation branches following the extractor.
While multi-task learning may improve both tasks, the SR and segmentation branches are independently trained.

Furthermore, CSSR in a paper proposed at a earlier conference~\cite{CSSR2021} showed that joint learning of SR can also improve crack segmentation. While the support by SR in previous methods, including CSSR, has been done with non-blind SR, we believe that making this support Blind SR based on joint learning will improve the robustness to blurred images presented in Problems $\langle D \rangle$ in Table~\ref{tab:addlabel}.


\section{Joint Learning of Blind SR and Crack Segmentation}
\label{section:method}

While methods using joint end-to-end learning with SR~\cite{DBLP:conf/iccv/SinghN0V19, DBLP:journals/tnn/ZhangBDXG21, bib:TDSR2021,DBLP:conf/cvpr/WangLZTS20} mentioned in Sec.~\ref{subsection:joint} are close to our work, it is difficult to apply them to crack segmentation in realistic scenarios.
This is because these methods using non-blind SR cannot cope with unknown blurs
observed in images degraded by out-of-focus and motion blurs.
Our CSBSR resolves this problem by employing blind SR in joint learning (Sec.~\ref{subsection:method_joint}). As far as we know, this is the first attempt at joint learning of blind SR. 
For further coping with the class-imbalance issue, this paper also proposes a new combination of loss functions for class-imbalance fine segmentation (Sec.~\ref{subsection:method_gbc_loss}).
In addition to joint learning, we propose loss weighting for optimizing segmentation more for SR in Sec.~\ref{subsection:method_intermediate} and extra skip connection paths for optimizing SR more for segmentation, as described in Secs.~\ref{subsection:method_skip}.

\subsection{Joint Learning}
\label{subsection:method_joint}

\begin{figure*}[t]
\begin{center}
\includegraphics[width=\textwidth]{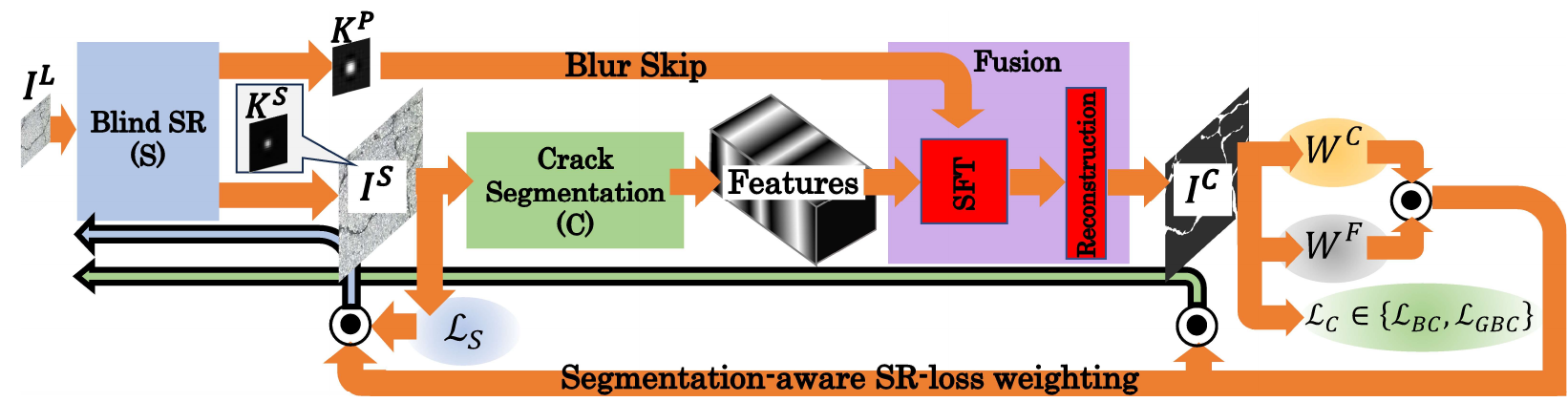}
\end{center}
\vspace{-1em}
\caption{Proposed joint learning network with blind SR and segmentation.
See the caption of Fig.~\ref{fig:joint_learning} for the explanations of arrows and ellipses. $\odot$ indicates a pixelwise multiplication operator.
While $K^{S}$ (i.e., blur kernel remaining in $I^{S}$) is unavailable and unused in our method, $K^{S}$ is shown for explanation of our blur skip scheme proposed in Sec.~\ref{subsection:method_skip}.}

\vspace{-1.5em}
\label{fig:network}
\end{figure*}

CSBSR consists of blind SR and segmentation networks, as shown in Fig.~\ref{fig:joint_learning} (c).
Its detail is shown in Fig.~\ref{fig:network}.
The blind SR network, $S$, maps $\boldsymbol{I}^{L}$ to its SR image $\boldsymbol{I}^{S}=S(\boldsymbol{I}^{L}) \in \mathbb{R}^{3\times h\times w}$.
The crack segmentation network, $C$, takes $\boldsymbol{I}^{S}$ and outputs a crack segmentation image $\boldsymbol{I}^{C}=C(\boldsymbol{I}^{S}) \in \mathbb{R}^{1\times h\times w}$.
Any differentiable SR and crack segmentation networks can be employed as $S$ and $C$, respectively.
Let $\mathcal{L}_{S} \in \mathbb{R}$ and $\mathcal{L}_{C} \in \mathbb{R}$ denote loss functions for $S$ and $C$, respectively.
The whole network is trained by the following loss $\mathcal{L}_{J} \in \mathbb{R}$ with the task weight $\beta~\in \mathbb{R}$ as a hyper-parameter where $0 \leq \beta \leq 1$:
\begin{eqnarray}
\mathcal{L}_{J} & = & (1 - \beta) \mathcal{L}_{S} + \beta \mathcal{L}_{C}
\label{eq:joint_loss}
\end{eqnarray}
Note that $\mathcal{L}_{S}$ and $\mathcal{L}_{C}$ are arbitrary loss functions for $S$ and $C$, respectively. $\mathcal{L}_{J}$ is defined as the weighted sum of $\mathcal{L}_{S}$ and $\mathcal{L}_{C}$. The effect of the weight, $\beta$, is empirically verified in Sec.~\ref{section:experiments}. The details of  $\mathcal{L}_{S}$ and $\mathcal{L}_{C}$ used in this paper are described in what follows.

\noindent {\bf Implementation details:}
In our experiments, Deep Back-Projection Network (DBPN)~\cite{bib:DBPN} and its extension to blind SR, which is called Kernelized Back-Projection Networks (KBPN)~\cite{bib:kbpn}, are employed as $S$ for fair comparison between our proposed methods with non-blind SR and blind SR (i.e., comparison between CSSR and CSBSR).
Different from DBPN as non-blind SR, KBPN also outputs its estimated blur kernel.
Loss functions used in DBPN and KBPN are used as $\mathcal{L}_{S}$ in our joint learning with no change.

$C$ is implemented with each of U-Net~\cite{bib:Ronneberger2015unet}, Pyramid Scene Parsing Network (PSPNet)~\cite{bib:zhao2017PSPNet}, CrackFormer~\cite{bib:liu2021crackformer}, and high-resolution network with Object-Contextual Representation (HRNet+OCR)~\cite{DBLP:conf/eccv/YuanCW20} for validating a wide applicability of our method.

Section~\ref{subsection:method_gbc_loss} proposes a new general-purpose segmentation loss, which is applicable to all of these networks as $\mathcal{L}_{C}$.

\subsection{Boundary Combo Loss}
\label{subsection:method_gbc_loss}

For suppressing class-imbalance difficulty in crack segmentation, we propose the BC loss that simultaneously achieves locally-fine and globally-robust segmentation.
Fine segmentation can be achieved by the boundary-based approach such as the Boundary loss~\cite{DBLP:journals/mia/KervadecBDGDA21}.
However, if only the boundary-based approach is employed, the segmentation network is easy to fall into local minima, as validated in~\cite{DBLP:journals/mia/KervadecBDGDA21}.
This problem can be resolved by employing the boundary-based approach simultaneously with a loss that evaluates the whole image region.
In~\cite{DBLP:journals/mia/KervadecBDGDA21}, the Generalized Dice (GDice) loss~\cite{DBLP:conf/miccai/SudreLVOC17} is empirically demonstrated to be a good choice.
%
However, it is reported that the Sigmoid function included in the GDice loss and its original Dice loss tends to cause the vanishing gradient problem~\cite{bib:taghanaki2019combo}.

This paper explores more appropriate losses combined with the Boundary loss for stable learning as well as fine segmentation.
We improve learning stability by combining the GDice loss with the WCE loss that is expressed without the derivative of the Sigmoid function, which tends to cause gradient vanishing.
Since the Dice loss and the WCE loss have different properties (i.e., which are categorized to region-based and distribution-based losses, respectively, as introduced in~\cite{DBLP:journals/mia/MaCNHLLYM21}), it is also validated that a pair of the Dice and WCE losses, which is called the Combo loss~\cite{bib:taghanaki2019combo}, complementarily work for better segmentation.
Finally, we propose the following BC loss $\mathcal{L}_{BC}$ $\in \mathbb{R}$, as $\mathcal{L}_{C}$ in our joint learning:
\begin{eqnarray}
\mathcal{L}_{BC}&=&\alpha\mathcal{L}_{B}+(1-\alpha)\left( (1-\gamma)\mathcal{L}_{D}+\gamma\mathcal{L}_{WCE} \right), \label{eq:bc}
\end{eqnarray}  
where $\mathcal{L}_{B}~\in \mathbb{R}, \mathcal{L}_{D}~\in \mathbb{R}$ and $\mathcal{L}_{WCE}~\in \mathbb{R}$ denote the Boundary~\cite{DBLP:journals/mia/KervadecBDGDA21}, Dice~\cite{bib:Dice}, and WCE~\cite{bib:WCE_crack} losses, respectively. 
$\alpha \in  \mathbb{R}$, where $0 \leq \alpha \leq 1$, and $\gamma \in \mathbb{R}$, where $0 \leq \gamma \leq 1$, are hyper-parameters.

In Eq.~(\ref{eq:bc}), first of all, $L_D$ and $L_{WCE}$ are summed with weight $\gamma$ because the effectiveness of this sum is verified in~\cite{bib:taghanaki2019combo}, and the appropriate value of $\gamma$ can be given by~\cite{bib:taghanaki2019combo}.
Next, this sum and $L_B$ is summed with weight $\alpha$.

The specific values of $\alpha$ and $\gamma$ are shown in Sec.~\ref{section:experiments}.
$\mathcal{L}_{BC}$ consists of the region, distribution, and boundary-based losses.
A combination of these three loss categories (i.e. $\mathcal{L}_{B}$ is boundary-based, $\mathcal{L}_{D}$ is region-based and $\mathcal{L}_{WCE}$ is distribution-based) are never evaluated according to the survey~\cite{DBLP:journals/mia/MaCNHLLYM21}.
As a variant of $\mathcal{L}_{BC}$, we also propose $\mathcal{L}_{GBC}~\in \mathbb{R}$~\cite{DBLP:conf/miccai/SudreLVOC17} in which the GDice loss $\mathcal{L}_{GD}~\in \mathbb{R}$ is used in $\mathcal{L}_{BC}$ instead of $\mathcal{L}_{D}$.


While one may refer to the original papers of $\mathcal{L}_{B}$, $\mathcal{L}_{D}$, $\mathcal{L}_{GD}$, and $\mathcal{L}_{WCE}$ for the details, these losses are briefly explained in the following three paragraphs.

\noindent {\bf Boundary loss ($\mathcal{L}_{B}$):}
The Boundary loss~\cite{DBLP:journals/mia/KervadecBDGDA21}, computes the distance-weighted 2D area between the GT crack and its estimated one, which becomes zero in the ideal estimation, as follows:
\begin{small}
\begin{eqnarray}
D(\partial G, \partial S)& \hspace*{-2mm} =&\int_{\partial G} ||y_{\partial S}(p)-p||^2 dp \nonumber \\
& \hspace*{-2mm} \simeq& 2 \int_{\varDelta S}D_G(q)dq \nonumber \\
& \hspace*{-2mm} =& 2 \bigl( \int_{\Omega}\phi_G(q)s(q)dq-\int_{\Omega}\phi_G(q)g(q) dq \bigr),
\end{eqnarray}
\end{small}
\hspace*{-3mm}
where $G$ and $S$ denote the pixel sets of the GT crack and its estimated one, respectively.
$p~\in \mathbb{R}$ and $y_{\partial S}(p)$ denote a point on boundary $\partial G$ and its corresponding point on boundary $\partial S$, respectively.$y_{\partial S}(p)$ is an intersection between $\partial S$ and a normal of $\partial G$ at $p$.
$\varDelta S = (S/G) \cup (G/S)$ is the mismatch part between $G$ and $S$.
$q \in \mathbb{R}$ donates a point in the image and $D_G(q)$ is the distance map defined by the length of the line segment when the normal of $\partial G$ intersects $q$.
$s(q)$ and $g(q)$ are binary indicator functions, where $s(q)=1$ and $g(q)=1$ if $q \in S$ and $q \in G$, respectively.
$\phi_G (q)$ is the level set representation of boundary $\partial G$: $\phi_G = -D_G(q)$ if $q \in G$, and $\phi_G = D_G(q)$ otherwise.
$\Omega$ denotes a pixel set in the image.
The second term in Eq.~(\ref{eq:Boundary}) is omitted as it is independent of the network parameters.
By replacing $s(q)$ by the network softmax outputs $s_\theta (q)$, we obtain the Boundary loss function below:
\begin{equation}
\mathcal{L}_{B} = \int_{\Omega} \phi_G (q) s_\theta (q) dq \label{eq:Boundary} 
\end{equation}

\noindent {\bf Dice and GDice losses ($\mathcal{L}_{D}$ and $\mathcal{L}_{GD}$):}
The Dice loss~\cite{bib:Dice} is a harmonic mean of precision and recall as expressed as follows:
\begin{eqnarray}
\mathcal{L}_{D}&=&\frac{2\sum^{M}_{j}\sum^{N}_{i}P_{ij} G_{ij}}{\sum^{M}_{j}\sum^{N}_{i}(P_{ij}^2+G_{ij}^2)}, \label{eq:dice}
\end{eqnarray}
where $M$ and $N$ denote the number of classes (i.e., $M=2$ in our problem) and the number of all pixels in each image, respectively.
$\boldsymbol{P} \in \mathbb{R}^{M\times N}$ and $\boldsymbol{G} \in \mathbb{R}^{M\times N}$ are the classification probability maps consisted of elements $P_{ij} \in \mathbb{R}$, where $0 \leq P_{ij} \leq 1$, and its GT consisted of elements $G_{ij} \in \mathbb{R}$, where $0 \leq G_{ij} \leq 1$.

Different from the Dice loss, the GDice loss~\cite{DBLP:conf/miccai/SudreLVOC17} is weighted by the number of pixels in each class as follows:
\begin{eqnarray}
\mathcal{L}_{GD}&=&\frac{2\sum^{M}_{j}w_{j}^{(GD)}\sum^{N}_{i}P_{ij} G_{ij}}{\sum^{M}_{j}w_{j}^{(GD)}\sum^{N}_{i}(P_{ij}+G_{ij})}, \label{eq:gdice}
\end{eqnarray}
where $\boldsymbol{w}^{(GD)} \in \mathbb{R}^{M} $ is weight coefficients per class consisted of elements  $w_{j}^{(GD)} = \frac{1}{\sum^{N}_{i}g_{ij}}~\in \mathbb{R}$.
\vspace{1mm}

\noindent {\bf WCE loss ($\mathcal{L}_{WCE}$):}
The WCE loss~\cite{bib:WCE_crack} is the Cross Entropy loss weighted by coefficients $\boldsymbol{w}^{(WCE)} \in \mathbb{R}^{M}$, which is consisted of elements $w_{j}^{(WCE)}=\frac{1}{\sum^{N'}_{i}g_{ij}}$ where $N' = N N_{I}~\in \mathbb{R}$ and $N_{I}~\in \mathbb{R}$ is the number of all training images:
\begin{eqnarray}
    \mathcal{L}_{WCE}= \sum^{M}_{j}w_{j}^{(WCE)}\sum^{N}_{i}G_{ij}\log P_{ij}
\label{eq:wce}
\end{eqnarray}


\subsection{Segmentation-aware Weights for SR}
\label{subsection:method_intermediate}

In addition to end-to-end learning with $\mathcal{L}_{C}$ (i.e., segmentation loss in Eq.~(\ref{eq:joint_loss})), we propose to weight $\mathcal{L}_{S}$ by $\mathcal{L}_{C}$ for further optimizing the SR network $S$ for segmentation.
This weighting is achieved by pixelwise multiplying $\mathcal{L}_{S}$ by $\mathcal{L}_{C}$.

It is not yet easy to discriminate between crack and background pixels for precisely detecting fine cracks.
This difficulty arises especially around crack pixels.
For such difficult pixelwise segmentation, our method employs the following two difficulty-aware weights:
\begin{itemize}
    \item For detecting all fine thin cracks, a segmentation loss function is weighted so that pixels inside cracks are weighted higher.
    A weight given to pixel $p$, $w^{C}_{p}~\in \mathbb{R}$, where $0 \leq w^{C}_{p} \leq 1$, is expressed as follows:
\begin{eqnarray}
w^{C}_{p} & = & \exp( - m^{C} D_{p} ) \label{eq:CO_weight}
\end{eqnarray}
where $m^{C}~\in \mathbb{R}$, where $0 < m^{C}$ and $D_{p}~\in \mathbb{R}$, where $0 \leq D_{p}$ denote a weight constant and a distance between $p$ and its nearest GT crack pixel, respectively. $w^{C}_{p}$ is called the Crack-Oriented (CO) weight.
    \item For hard pixel mining, a segmentation loss function is weighted so that eroneous pixels are weighted higher.
    The erroneous pixel is defined so that the difference between the prediction (i.e., $T^{P}_{p}$) and its ground-truth (i.e., $T^{GT}_{p}$) is higher in the erroneous pixel.
    For such difficulty-aware segmentation, in our method, a weight given to pixel $p$, $w^{F}_{p}~\in \mathbb{R}$, where $1 \leq w^{F}_{p}$, is expressed as follows:
    \begin{eqnarray}
    w^{F}_{p} & = & \exp( m^{F} | T^{P}_{p} - T^{GT}_{p} | ),
    \label{eq:FO_weight}
\end{eqnarray}
where $T^{P}_{p} \in \mathbb{R}$, where $0 \leq T^{P}_{p} \leq 1$ and $T^{GT}_{p} \in \mathbb{R}$, where $T^{GT}_{p}  \in \{0, 1\}$ denote the value of $p$-th pixel in predicted and GT segmentation images, respectively.
$m^{F}~\in \mathbb{R}$, where $0 < m^{F}$ is a weight constant.
Our $w^{F}_{p}$ is applicable to any loss function such as our BC loss, Eq. (\ref{eq:bc}), consisting of multiple loss functions, while the focal loss~\cite{DBLP:conf/iccv/LinGGHD17} and the anchor loss~\cite{DBLP:conf/iccv/RyouJP19}, both of which also penalize hard samples, are based on a weighted cross entropy loss.
$w^{F}_{p}$ is called the Fail-Oriented (FO) weight.
\end{itemize}

\noindent
These two weights (\ref{eq:CO_weight}) and (\ref{eq:FO_weight}) are multiplied pixelwise by $\mathcal{L}_{S}$. The specific values of $m^C$ and $m^F$ are verified in section~\ref{subsubsec:element-verification}.

\subsection{Blur Skip for Blur-reflected Task Learning}
\label{subsection:method_skip}

It is not easy for the blind SR network to perfectly predict the GT blur kernel $\boldsymbol{K}$ and the GT HR image $\boldsymbol{I}^{H}$ so that $\boldsymbol{I}^{S} = \boldsymbol{I}^{H}$.
Let $\boldsymbol{K}^{P} \in \mathbb{R}^{k\times k}$ and $\boldsymbol{K}^{S} \in \mathbb{R}^{k\times k}$ denote the predicted kernel and the blur kernel that remains in $\boldsymbol{I}^{S}$ so that $\boldsymbol{K} = \boldsymbol{K}^{P} + \boldsymbol{K}^{S}$ and $\\boldsymbol{I}^{S} = \boldsymbol{I}^{H} \ast \boldsymbol{K}^{S}$.
We assume that $\boldsymbol{K}^{S}$ correlates with $\boldsymbol{K}^{P}$.

\begin{figure*}[t]
    \begin{center}
    \includegraphics[width=\linewidth]{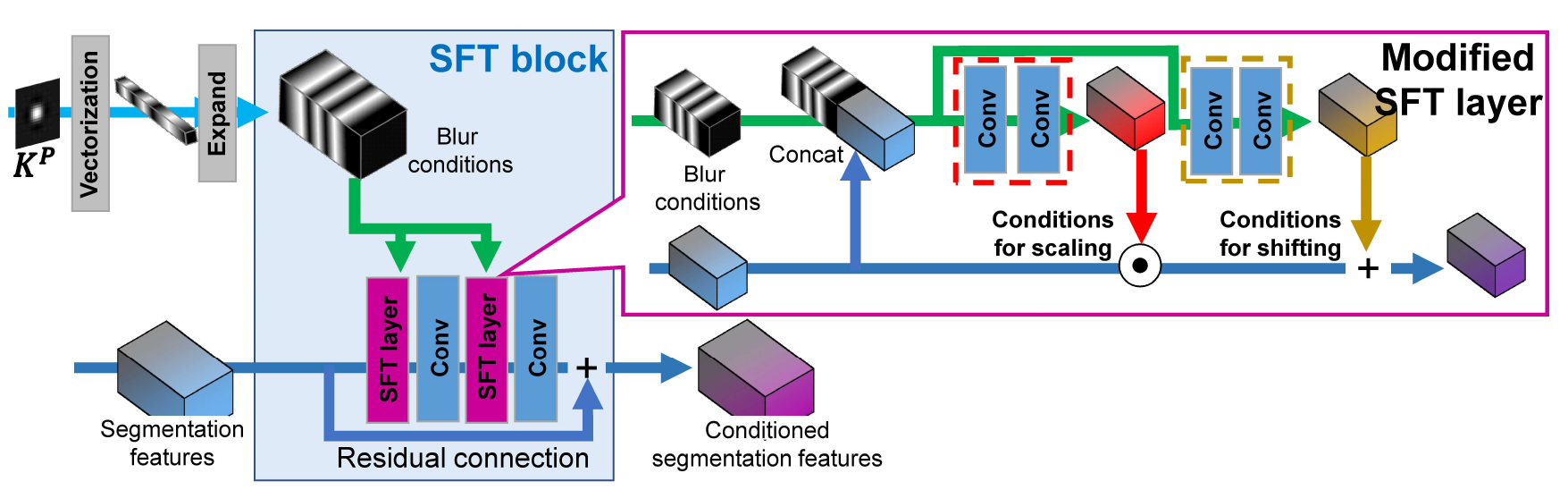}
    \vspace{-2em}
    \end{center}
    \caption{
    The structure of our blur skip module using SFT~\cite{DBLP:conf/cvpr/WangYDL18}.
    Each 3D box and rectangle depict a feature set and a process, respectively. $\odot$ indicates a pixelwise multiplication operator. Conv means convolution layer.
    }
    \label{fig:blurskip}
\end{figure*}

Based on this assumption, this paper proposes blur-reflected segmentation learning via a skip connection, which is called the blur skip, from the SR network $S$ to the segmentation network $C$.
This skip connection forwards $\boldsymbol{K}^{P}$ to the end of $C$ in order to condition features extracted by $C$ with $\boldsymbol{K}^{P}$. 
While this conditioning is achieved by the Spatial Feature Transform (SFT) [54], SFT is marginally modified for CSBSR as follows. The detail of the modified SFT layer is shown in Fig.~\ref{fig:blurskip}. In the original SFT layer, conditions are directly fed into convolution (conv) layers for producing conditioning features (which are depicted by red and yellow 3D boxes, respectively, in Fig.~\ref{fig:blurskip}) for scaling and shifting. Different from this original SFT layer, target features (``Segmentation features'' in Fig.~\ref{fig:blurskip}) are concatenated to the conditions.
When using our modified SFT for blur skip compared to the original SFT, under the conditions in the bottom row of Table~\ref{table:exp_kernel_skip}, the maximum value of Intersect of Union (IoU) increases by 0.035, and the minimum value of 95\% Hausdorff Distance (HD95)~\cite{bib:code_hd95} decrease by 17.12. We confirmed that our modified SFT contributed to improving segmentation performance. For more details on the IoU and HD95 metrics, please refer to ``Evaluation Metrics'' of Sec.~\ref{subsection:exp_synthetic}.

\subsection{Training Strategy}
\label{subsection:strategy}

Our joint learning has several loss functions, weights, and hyper-parameters.
They should be properly used for training our complex network consisting of $S$ and $C$.

\begin{description}
    \item[Step 1:] As with most tasks each of which has a limited amount of training data,
    $S$ is pre-trained with general huge datasets for blind SR.
    \item[Step 2:] With a dataset for crack segmentation, only $S$ is initially finetuned with $\beta=0$ in Eq~.(\ref{eq:joint_loss}).
    \item[Step 3:] The whole network is finetuned so that $C$ is weighted by a constant (i.e., $\beta \neq 0$).
\end{description}

\section{Experimental Results}
\label{section:experiments}

\subsection{Pre-training and Training Details}
\label{subsection:exp_detail}

For pre-training the SR network $S$, \num{3450} images in the DIVerse 2K resolution image (DIV2K) dataset~\cite{bib:ntire2017} (800 images) and the Flickr 2K resolution image (Flickr2K) dataset~\cite{bib:Flickr2K} (\num{2650} images) were used.
The whole network for crack segmentation $C$ was not pre-trained but its feature extractor was pre-trained with the ImageNet~\cite{bib:deng2009imagenet}.


For pre-training $S$ (i.e., Step 1 in Sec.~\ref{subsection:strategy}) and finetuning $S$ and $C$ (i.e., Steps 2 and 3), an image patch fed into each network is randomly cropped with vertical and horizontal flips from each training image for data augmentation.
This patch is regarded as a HR image $\boldsymbol{I}^{H}$.
From $\boldsymbol{I}^{H}$, its LR images $\boldsymbol{I}^{L}$ are generated with various blur kernels $\boldsymbol{K}$ and bicubic downsampling $\downarrow_{s}$, as expressed in Eq. (\ref{eq:down_model}).
$\boldsymbol{K}$ is randomly sampled from the anisotropic 2D Gaussian blurs with variance $\sigma_a^2$$\in \mathbb{R}$, where $0.2 \leq \sigma_a^2 \leq 4.0$, $\sigma_b^2 \in \mathbb{R}$, where $0.2 \leq \sigma_b^2 \leq 4.0$ and angle $\theta_{gaus}\in\mathbb{R}$, where $0\leq \theta_{gaus} < \pi$.
The kernel size $k \times k$ is $21 \times 21$ pixels.
The scaling factor $s$ is 4.
The feature extractor of $C$ is pre-trained depending on the segmentation network as follows.
For U-Net and PSPNet, VGG-16~\cite{DBLP:journals/corr/SimonyanZ14a} is provided by torchvision~\cite{bib:torchvision}.
For HRNet+OCR, the author's model~\cite{DBLP:conf/eccv/YuanCW20} is used.

For pre-training of $S$ in Step 1, the number of iterations is \num{200000}.
The minibatch size is six.
The segmentation model used here includes Batch Normalization, so the batch size is one of the important parameters. With limited computational resources, however, we have to determine a good trade-off between the batch size and the input image size. Since our problem is crack segmentation in low-resolution images, the image size should not be further reduced. In addition, the global structure of the cracks is useful for detecting thin cracks. Therefore, the input image should be fed into our joint network without cropping small windows from the input image. In our experiments, therefore, the batch size is maximized under the condition that the input image size is not changed.

Adam~\cite{Adam} is used as an optimizer with $\beta_1=0.9, \beta_2=0.999, \epsilon=10^{-8}$.
The learning rate is $2\times 10^{-4}$. The number of iterations is \num{30000} and \num{150000} in Steps 2 and 3, respectively.
The minibatch size and the optimizer are equal to those in the aforementioned pre-training.
The learning rate is $2\times 10^{-5}$. All experiments were performed with one to a maximum of six NVIDIA A100 GPUs and two AMD EPYC 7302 CPUs.

\subsection{Synthetically-degraded Crack Images}
\label{subsection:exp_synthetic}

\begin{figure*}[t]
\begin{center}
\includegraphics[width=\textwidth]{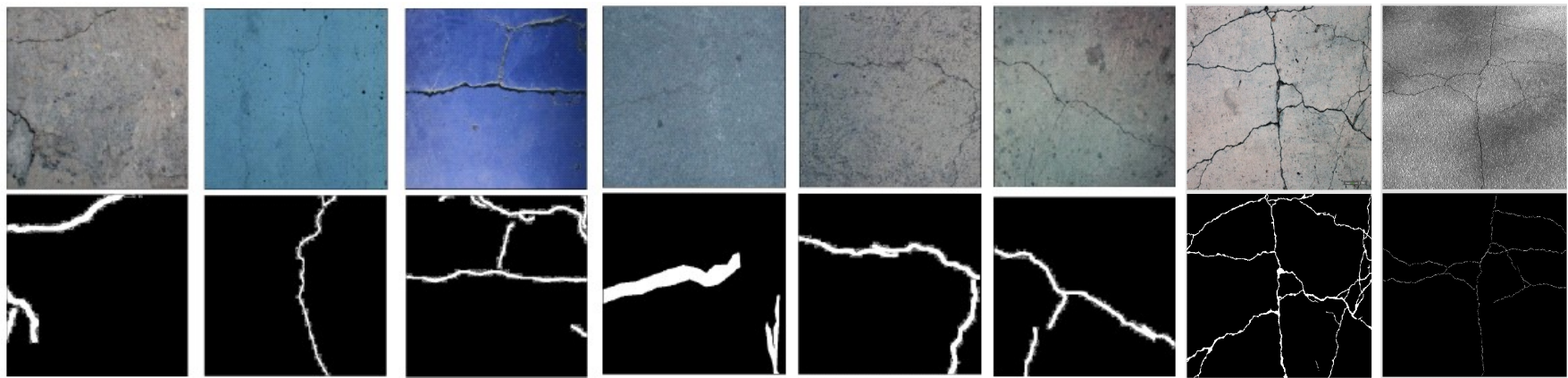}
\end{center}
\vspace{-1em}
\caption{Sample images in the Khanhha dataset~\cite{bib:Khanhha}. The top row is the RGB image treated as $\boldsymbol{I}^H$ in this paper, and the bottom row is the GT of the segmentation.}
\vspace{-1em}
\label{fig:sample_images}
\end{figure*}

\noindent {\bf Training:}
For experiments shown in Secs.~\ref{subsection:exp_synthetic} and~\ref{subsection:exp_real},
the Khanhha dataset~\cite{bib:Khanhha} was used to finetune the whole network for CSBSR.
the SR and segmentation networks.
This dataset consists of CRACK500~\cite{bib:zhang2016road}, German Asphalt Pavement distress (GAPs)~\cite{bib:eisenbach2017how}, CrackForest~\cite{bib:shi2016automatic}, three data named Aigle-RN, ESAR and LCMS (AEL)~\cite{bib:amhaz2016automatic},  cracktree200~\cite{bib:zou2012cracktree}, DeepCrack~\cite{bib:liu2019deepcrack}, and Concrete Structure Spalling and Crack (CSSR)~\cite{bib:CSSC} datasets.
As shown in the sample images of these datasets, (Fig.~\ref{fig:sample_images}), the Khanhha dataset is challenging so that a variety of structures are observed and the properties of annotated cracks differ between the elemental datasets~\cite{bib:zhang2016road,bib:eisenbach2017how,bib:shi2016automatic,bib:amhaz2016automatic,bib:zou2012cracktree,bib:liu2019deepcrack,bib:CSSC}.
In the Khanhha dataset, the image size is $448 \times 448$ pixels, which is regarded as a HR image in our experiments.
The dataset has \num{9122481}, and \num{1695} training, validation, and test images.
These training and test sets were used as training images for all experiments and test images in experiments shown in Sec.~\ref{subsection:exp_synthetic}, respectively.

\noindent {\bf Evaluation Metrics:} 
Each SR image is evaluated with Peak Signal-to-Noise Ratio (PSNR) and Structural Similarity Index Measure (SSIM)~\cite{bib:SSIM}.
Each segmentation image is evaluated with IoU.
While IoU is computed in a binarized image, the output of CSBSR is a segmentation image in which each pixel has a probability of being a crack or not a crack.
Since IoU differs depending on a threshold for binarization, the threshold for each method is determined so that the mean IoU over all test images is maximized.
This maximized IoU is called IoU$_{max}$.
For evaluation independently of thresholding, IoUs are averaged over thresholds (AIU)~\cite{bib:Yang2020}.
While IoU is a major metric for segmentation, it is inappropriate for evaluating fine thin cracks because a slight displacement makes IoU significantly small even if the structures of GT and estimated cracks are almost similar.
For appropriately evaluating such similar cracks,HD95~\cite{bib:code_hd95} is employed.
As with IoU, the HD95 threshold for each method is also determined so that the mean HD95 over all test images is minimized.
This minimized HD95 is called HD95$_{min}$.
For evaluation independently of thresholding, HD95s are also averaged over thresholds.
This averaged HD95 is called AHD95.

\subsubsection{Comparison with SoTA segmentation methods}\hfill\\
For comparative experiments, 1,695 HR test images in the Khanhha dataset are degraded to their LR images in the same manner as training image generation.

For validating the wide applicability of CSBSR, four SoTA segmentation networks (i.e., PSPNet~\cite{bib:zhao2017PSPNet} for Table\footnote{In this paper, w/o means without and w/ means with in the figures and tables.}~\ref{table:exp_sota} (e), HRNet+OCR~\cite{DBLP:conf/eccv/YuanCW20} for Table~\ref{table:exp_sota} (g), CrackFormer~\cite{bib:liu2021crackformer} for Table~\ref{table:exp_sota} (i), and U-Net~\cite{bib:Ronneberger2015unet} for Table~\ref{table:exp_sota} (k)) are used as a segmentation network in CSBSR, as described in Sec.~\ref{subsection:method_joint}.
While CSBSR is trained in a joint end-to-end manner (i.e., (e), (g), (i), (k) in Table~\ref{table:exp_sota}), the results of independent blind SR and segmentation networks (i.e., (d), (f), (h), (j) in Table~\ref{table:exp_sota}) are also shown for comparison.
To focus on the difference between the network architectures for segmentation, all of these segmentation networks are trained with our BC loss in Eq.~(\ref{eq:bc}).
In BC loss, $\alpha$ is dynamically determined during the learning phese by the rebalance strategy of the previous study~\cite{DBLP:journals/mia/KervadecBDGDA21} and $\gamma=0.5$ follows the previous study~\cite{bib:taghanaki2019combo}.
The task weight $\beta$ in Eq. (\ref{eq:joint_loss}) is determined empirically for each method and fixed during Step 3 in the training strategy (Sec.~\ref{subsection:strategy}).

In addition, CSBSR is compared with SoTA methods in which non-blind SR and segmentation are used (i.e.,  Table~\ref{table:exp_sota} (b) Deep Super Resolution Crack network (SrcNet)~\cite{bib:SrcNet} in which SR and segmentation are trained independently and  Table~\ref{table:exp_sota} (c) DSRL~\cite{DBLP:conf/cvpr/WangLZTS20} in which SR and segmentation are trained in a multi-task learning manner).
The segmentation network of SrcNet and DSRL is trained with the Binary Cross Entropy loss.
While SrcNet is implemented by ourselves because its code is not available, we used the publicly-available implementation of DSRL~\cite{DBLP:conf/cvpr/WangLZTS20}.

\begin{table*}[t]
  \centering
  \caption{
  Khanhha dataset results compared with CSBSR using different segmentation networks. Joint learning's impact validated by training SR and segmentation networks separately in (d - k). SoTA method's results also compared in (b) and (c). High-resolution image used directly for segmentation in one case for upper bound analysis in (a). Best scores highlighted in \red{red}.}
  \label{table:exp_sota}
    \small
    \begin{tabular}{lcccccc} 
    \toprule 
    {} & \multicolumn{4}{c}{Segmentation metrics} & \multicolumn{2}{c}{SR metrics} \\
    {Model} & {IoU$_{max}\uparrow$} & {~AIU$\uparrow$} & {~HD95$_{min}\downarrow$} & {~AHD95$\downarrow$} & {~PSNR$\uparrow$} & {~SSIM$\uparrow$}\\ 
    \cline{1-7}
    \midrule
    (a) Segmentation in HR & 0.616 & 0.559 & 6.20 & 11.89 & - & - \\
    \cline{1-7}
    \midrule
    (b) SrcNet~\cite{bib:SrcNet} & 0.368 & 0.320 & 95.16 & 130.47 & 27.82 & 0.639 \\
    (c) DSRL~\cite{DBLP:conf/cvpr/WangLZTS20} & 0.391 & 0.285 & 44.23 & 148.97 & 20.16 & 0.501 \\ 
    \cline{1-7}
    \midrule
    (d) KBPN + PSPNet~\cite{bib:zhao2017PSPNet} & 0.548 & 0.524 & 28.45 & 31.62 & 28.62 & 0.706 \\
    (e) CSBSR w/~\cite{bib:zhao2017PSPNet} ($\beta=0.3$) & \textcolor{red}{\textbf{0.573}} & \textcolor{red}{\textbf{0.552}} & 20.92 & 22.52 & \textcolor{red}{\textbf{28.75}} & 0.703 \\
    \cline{1-7}
    (f) KBPN + (HRNet+OCR~\cite{DBLP:conf/eccv/YuanCW20}) & 0.522 & 0.501 & 26.45 & 28.74 & 28.68 & 0.706 \\ 
    (g) CSBSR w/~\cite{DBLP:conf/eccv/YuanCW20} ($\beta=0.9$) & 0.553 & 0.534 & \textcolor{red}{\textbf{17.54}} & \textcolor{red}{\textbf{20.29}} & 27.66 & 0.668 \\
    \cline{1-7}
    (h) KBPN + CrackFormer~\cite{bib:liu2021crackformer} & 0.447 & 0.424 & 46.86 & 58.91 & 28.68 & \textcolor{red}{\textbf{0.706}} \\ 
    (i) CSBSR w/~\cite{bib:liu2021crackformer} ($\beta=0.9$) & 0.469 & 0.443 & 39.37 & 56.59 & 25.93 & 0.571 \\
    \cline{1-7}
    (j) KBPN + U-Net~\cite{bib:Ronneberger2015unet} & 0.470 & 0.455 & 45.26 & 45.94 & 28.68 & 0.706 \\ 
    (k) CSBSR w/~\cite{bib:Ronneberger2015unet} ($\beta=0.3$) & 0.530 & 0.506 & 26.33 & 27.24 & 28.68 & 0.702 \\
    \bottomrule
    \end{tabular}
\end{table*}

\noindent
{\bf Quantitative Results:}
Table~\ref{table:exp_sota} shows quantitative results.
In all metrics, all variants of CSBSR are better than their original 
segmentation
methods.
That is, (e), (g), (i), and (k) are better than (d), (f), (h), and (j), respectively, in  Table~\ref{table:exp_sota}.
As a result, CSBSR is the best in all segmentation metrics (i.e., IoU, AIU, HD95, and AHD95).

Our proposed methods are also compared with SoTA segmentation methods using SR (i.e., (b) and (c) in  Table~\ref{table:exp_sota}).
The performance improvement of CSBSR compared to SrcNet might be acquired by BC loss, joint learning, and/or blind SR.
In comparison between CSBSR and DSRL, we can see the effectiveness of serial joint learning, as well as BC loss and blind SR.

Even in comparison with (a) segmentation in HR images
(implemented by PSPNet with BC loss),
the segmentation scores of CSBSR get close to those of segmentation in HR.
For example, IoU and AIU of CSBSR with PSPNet are 93.0\% and 98.7\% of those of segmentation in HR.
In terms of HD95, on the other hand, CSBSR is much inferior to segmentation in HR.
This reveals that CSBSR should be improved more in order to extract fine crack structures.

\begin{figure}[ht!]
        \begin{center}
        \hspace{-5mm}
        \includegraphics[width=\linewidth]{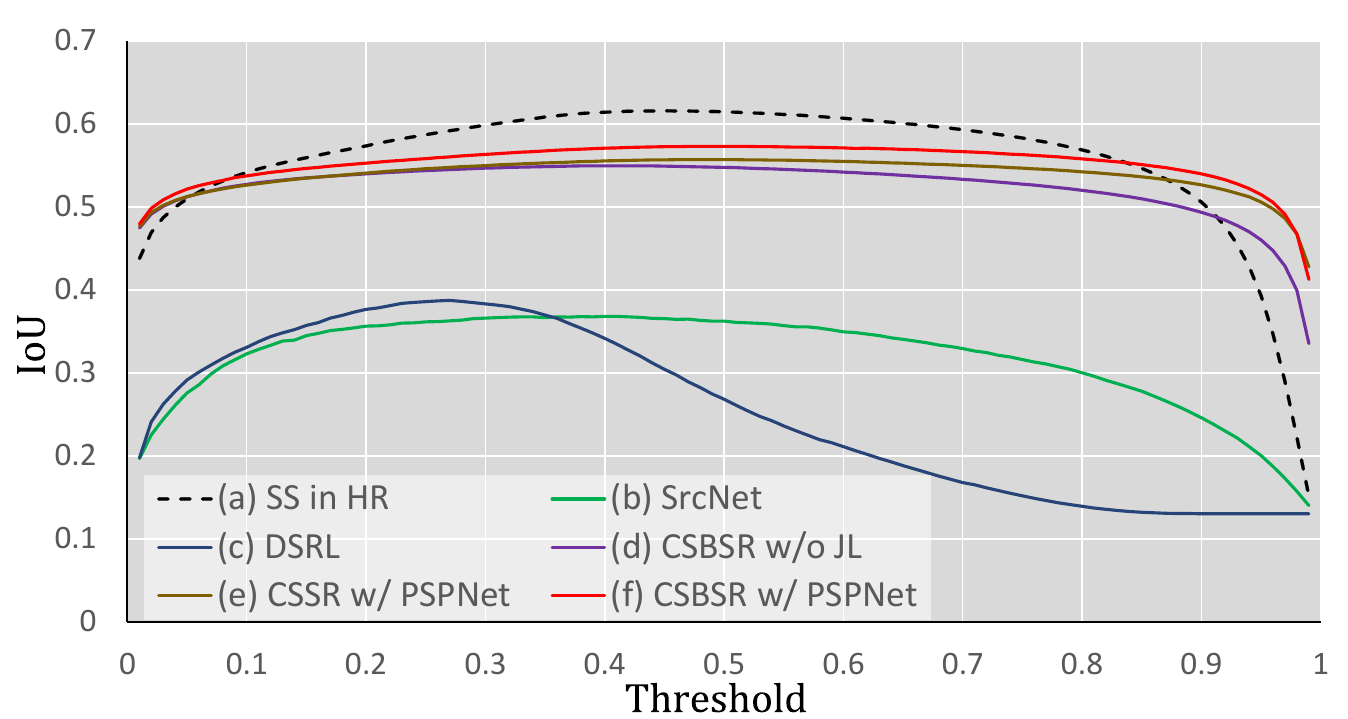}
        \end{center}
        \vspace{-1em}
        \caption{
        IoU comparison with SoTA methods on the Khanhha dataset. (a) HR segmentation by PSPNet. (b) SR segmentation by SrcNet. (c) SR segmentation by DSRL. (d) SR segmentation by CSBSR without joint learning. (e) SR segmentation by CSSR. (f) SR segmentation by CSBSR.
        }
        \label{fig:sota_iou}
        \begin{center}
        \hspace{-6mm}
        \includegraphics[width=\linewidth]{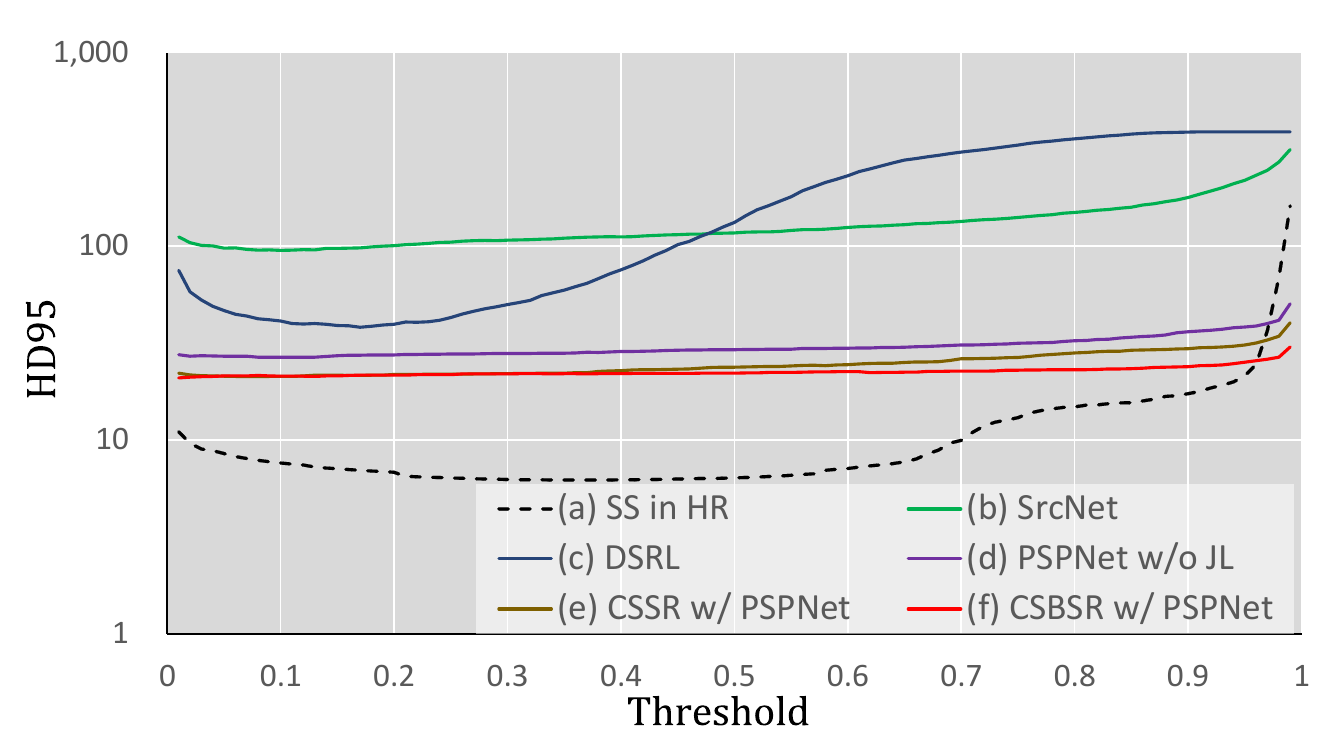}
        \end{center}
        \vspace{-1em}
        \caption{
        HD95 comparison with SoTA methods on the Khanhha dataset. \Erase{(a) HR segmentation by PSPNet. (b) SR segmentation by SrcNet. (c) SR segmentation by DSRL. (d) SR segmentation by CSBSR without joint learning. (e) SR segmentation by CSSR. (f) SR segmentation by CSBSR.}}
        \label{fig:sota_hd}
\end{figure}

The IoU and HD95 scores of our proposed method with CSBSR are shown in Fig.~\ref{fig:sota_iou} and Fig.~\ref{fig:sota_hd}.
For comparison, our method with non-blind SR (i.e., CSSR) and SoTA segmentation methods using SR are compared with CSBSR.
As the upper limitation, the scores of segmentation on GT HR images are also shown as (a) the black dashed lines in Fig.~\ref{fig:sota_iou} and Fig.~\ref{fig:sota_hd}, while LR images are fed into all other methods (b), (c), (d), (e), and (f) in Fig.~\ref{fig:sota_iou} and Fig.~\ref{fig:sota_hd}. It can be seen that (b) SrcNet and (c) DSRL are clearly inferior to others in both IoU and HD95.
In particular, the scores of DSRL are significantly changed depending on a change in the threshold.
This reveals that DSRL is sensitive to a change in the threshold.
The scores of all other methods accepting LR images are close to those of (a) segmentation in HR images.
In particular, Fig.~\ref{fig:sota_iou} (f) and Fig.~\ref{fig:sota_hd} (f) CSBSR can get higher scores in a wide range of the thresholds.
This stability against a change in the threshold is crucial in applying CSBSR to a variety of segmentation tasks.

\begin{figure*}[t]
\begin{center}
\includegraphics[width=0.8\textwidth]{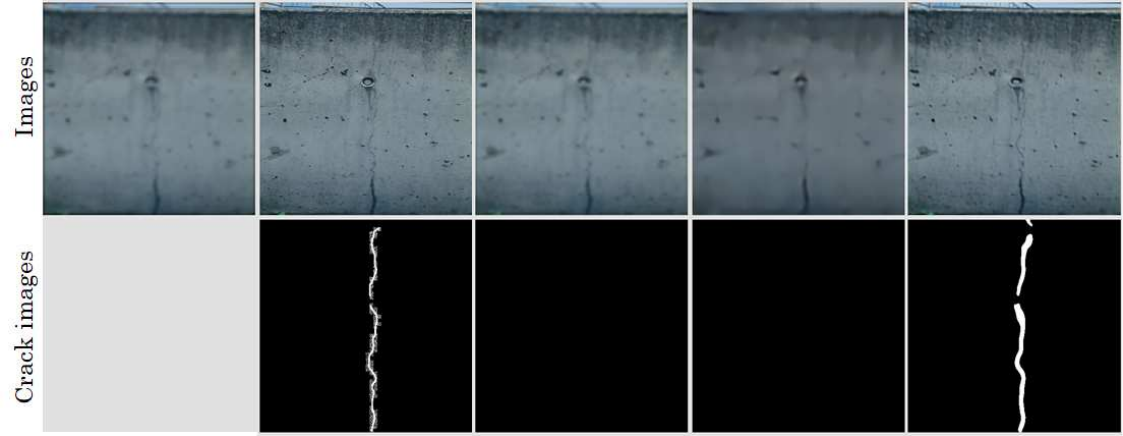}\\
~\hspace*{-1mm} (a) Input LR \hspace*{8mm} (b) HR GT  \hspace*{8mm}
(c) SrcNet~\cite{bib:SrcNet} \hspace*{7mm}
(d) DSRL~\cite{DBLP:conf/cvpr/WangLZTS20} \hspace*{8mm}
(e) CSBSR
\end{center}
\vspace{-0.5em}
\caption{Visual results of comparative experiments on the Khanhha dataset. In the upper row: (a) Input LR image (enlarged by Bicubic interpolation for visualization). (b) GT HR image. (c) SR image obtained by SrcNet. (d) SR image obtained by DSRL. (e) SR image obtained by our CSBSR. In the lower row of each example: (a) No image. (b) GT segmentation image in HR. (c) SR segmentation image obtained by SrcNet. (d) SR segmentation image obtained by DSRL. (e) SR segmentation image obtained by our CSBSR.}
\label{fig:visual_results_sota}
\vspace{-1em}
\end{figure*}

\noindent
{\bf Visual Results:}
Fig.~\ref{fig:visual_results_sota} shows visual results.
In the upper row, from left to right, the first and second images are an input LR image (enlarged by nearest neighbor interpolation) and its GT HR image.
The remaining three images are SR images of SrcNet, DSRL, and CSBSR.
It can be seen that the SR image of CSBSR is much sharper than those of SrcNet and DSRL.
In terms of the crack segmentation image also, CSBSR outperforms SrcNet and DSRL.

\begin{figure*}[ht!]
        \begin{center}
        \includegraphics[width=0.9\textwidth]{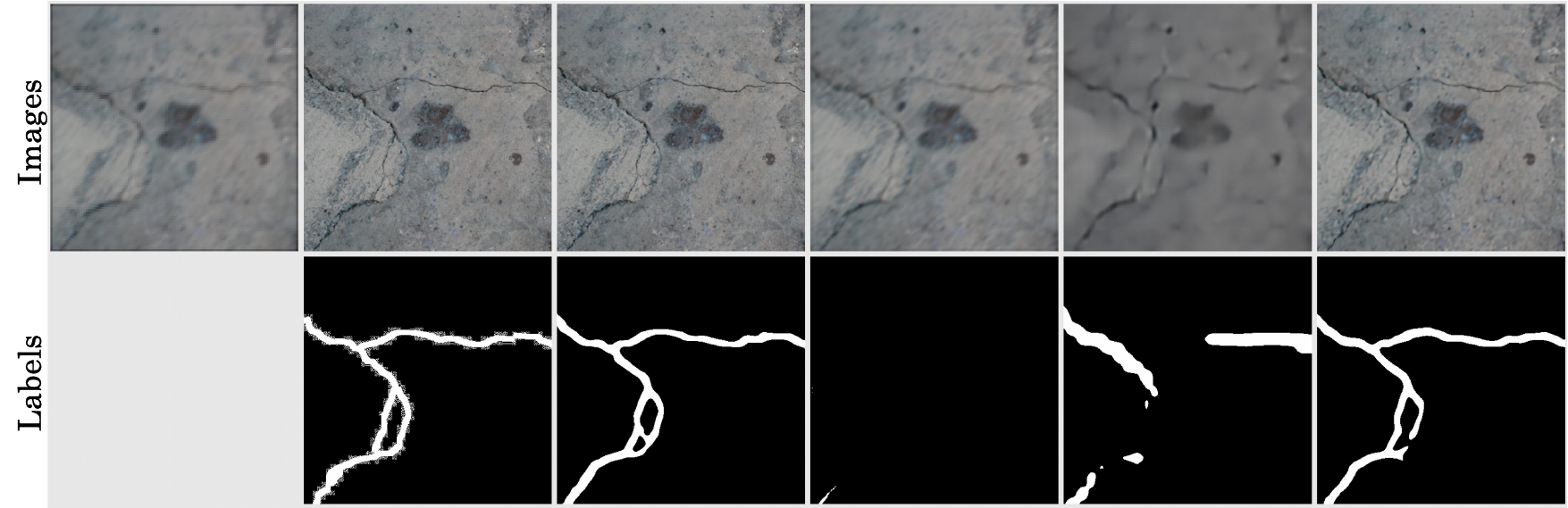}\\
        \includegraphics[width=0.9\textwidth]{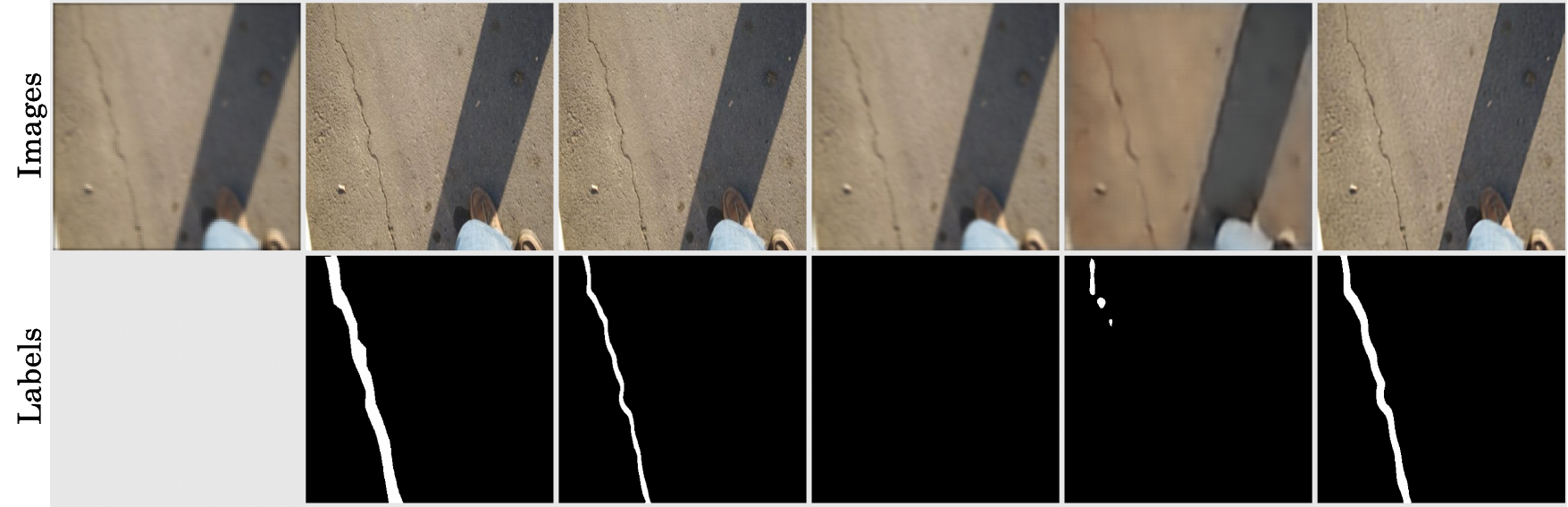}\\
        ~~~
        ~\hspace*{-2mm} (a) Input LR \hspace*{6mm} (b) HR GT  \hspace*{6mm} 
        (c) SS in HR \hspace*{5mm}
        (d) SrcNet~\cite{bib:SrcNet} \hspace*{4mm}
        (e) DSRL~\cite{DBLP:conf/cvpr/WangLZTS20} \hspace*{5mm}
        (f) CSBSR
        \end{center}
        \vspace{-1em}
        \caption{
        Visual comparison on the Khanhha dataset.
        \Replace{In the upper row of each example: (a) Input LR image (enlarged by Bicubic interpolation for visualization). (b, c) GT HR image. (d) SR image obtained by SrcNet. (e) SR image obtained by DSRL. (f) SR image obtained by our CSBSR.
        In the lower row of each example: (a) No image.
        (b) GT segmentation image in HR. (c) HR segmentation image obtained by PSPNet (d) SR segmentation image obtained by SrcNet. (e) SR segmentation image obtained by DSRL. (f) SR segmentation image obtained by our CSBSR.} (c) Upper row for each sample: GT HR image, Lower row for each sample: HR segmentation image obtained by PSPNet. 
        }
        \label{fig:crack1}
\end{figure*}

Fig.~\ref{fig:crack1} shows the examples of more complex cracks.
Since such complex crack pixels make it difficult to correctly detect these pixels, even segmentation methods using SR reconstruction (i.e., SrcNet~\cite{bib:SrcNet} and DSRL~\cite{DBLP:conf/cvpr/WangLZTS20}) cannot detect many crack pixels, as shown in Fig.~\ref{fig:crack1} (d) and (e).
As shown in Fig.~\ref{fig:crack1} (f), on the other hand, our CSBSR can obtain crack segmentation images that are similar to their corresponding segmentation images obtained in the original HR images shown in Fig.~\ref{fig:crack1} (c).
It can also be seen that CSBSR can reconstruct and detect even thin fine cracks in the SR image and segmentation images, respectively.
As a result, our results are similar to the GT segmentation images shown in Fig.~\ref{fig:crack1} (b).

\begin{figure*}[!t]
        \begin{center}
        \includegraphics[width=0.9\textwidth]{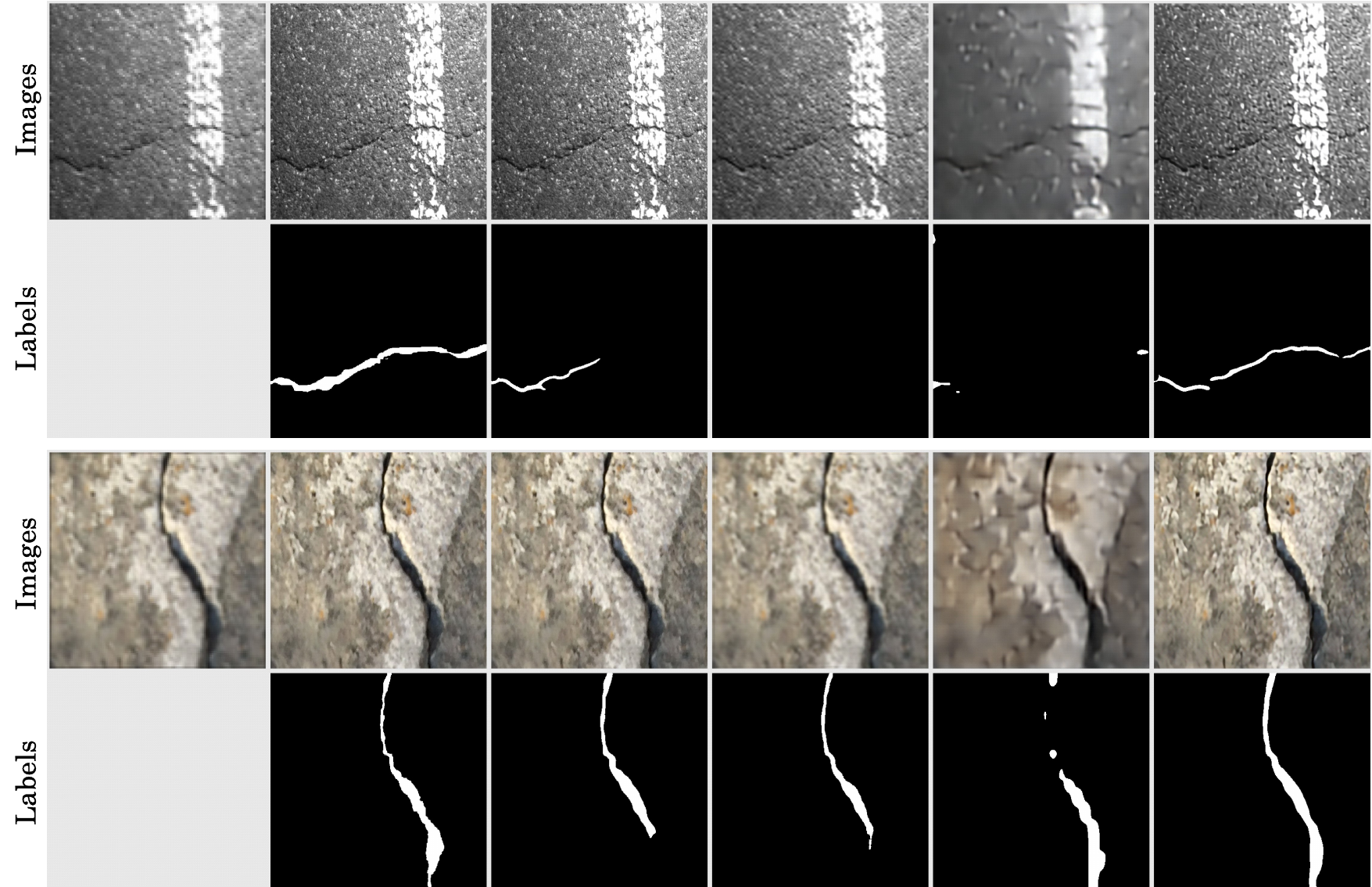}\\
        ~~~
        ~\hspace*{-3mm} (a) Input LR \hspace*{7mm} (b) HR GT  \hspace*{6mm} 
        (c) SS in HR \hspace*{5mm}
        (d) SrcNet~\cite{bib:SrcNet} \hspace*{4mm}
        (e) DSRL~\cite{DBLP:conf/cvpr/WangLZTS20} \hspace*{6mm}
        (f) CSBSR
        \end{center}
        \vspace{-1em}
        \caption{
        Examples where (f) the SR segmentation image obtained by our CSBSR is better than (c) the HR segmentation image obtained in the GT HR image.
        \Erase{In the upper row of each example: (a) Input LR image (enlarged by Bicubic interpolation for visualization). (b, c) GT HR image. (d) SR image obtained by SrcNet. (e) SR image obtained by DSRL. (f) SR image obtained by our CSBSR.
        In the lower row of each example: (a) No image.
        (b) GT segmentation image in HR. (c) HR segmentation image obtained by PSPNet (d) SR segmentation image obtained by SrcNet. (e) SR segmentation image obtained by DSRL. (f) SR segmentation image obtained by our CSBSR.}
        }
        \label{fig:crack2}
        \vspace{-1.5em}
\end{figure*}

Fig.~\ref{fig:crack2} shows examples where (f) the SR segmentation image obtained by CSBSR is better even than (c) the HR segmentation image obtained in the GT HR image.
These images are characterized by low image-contrast around crack pixels, thin cracks, and/or local illumination change around crack pixels.

We interpret that, while it is difficult for SR to reconstruct and for segmentation to detect such high-frequency structures and low-contrast structures shown in Figs.~\ref{fig:crack1} and~\ref{fig:crack2}, our joint learning of SR and segmentation with the segmentation-aware SR loss and the blur skip for blur-reflected segmentation learning can achieve these difficult tasks.

\begin{figure*}[ht!]
        \begin{center}
        \includegraphics[width=0.9\textwidth,trim={0 13.4mm 0 0},clip]{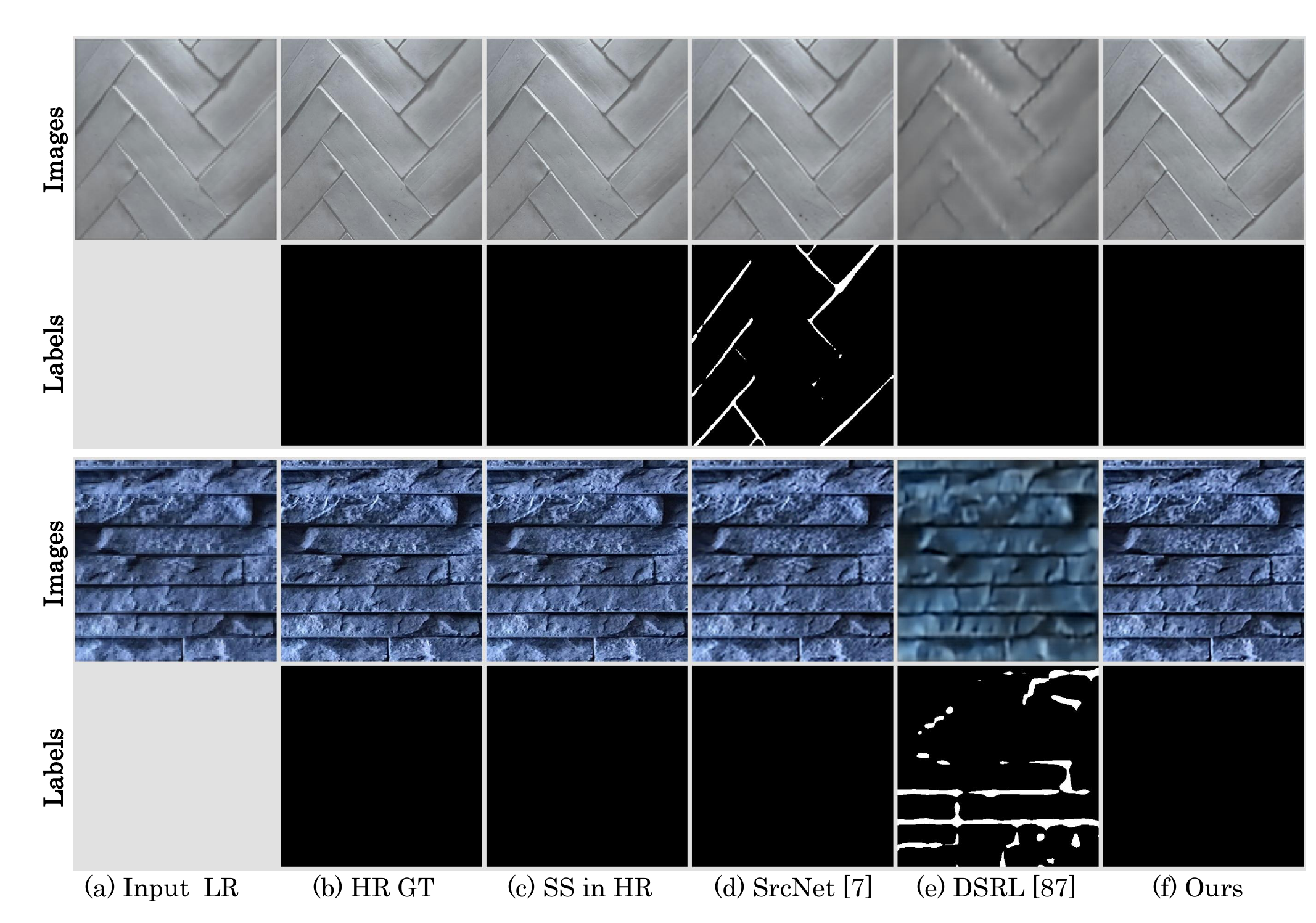}\\
        ~~~
        ~\hspace*{-2mm} (a) Input LR \hspace*{5mm} (b) HR GT  \hspace*{5mm} 
        (c) SS in HR \hspace*{5mm}
        (d) SrcNet~\cite{bib:SrcNet} \hspace*{3mm}
        (e) DSRL~\cite{DBLP:conf/cvpr/WangLZTS20} \hspace*{5mm}
        (f) CSBSR
        \end{center}
        \vspace{-1em}
        \caption{
        Examples where there are no crack pixels in (a) input LR image.
        \Erase{In the upper row of each example: (a) Input LR image (enlarged by Bicubic interpolation for visualization). (b, c) GT HR image. (d) SR image obtained by SrcNet. (e) SR image obtained by DSRL. (f) SR image obtained by our CSBSR.
        In the lower row of each example: (a) No image.
        (b) GT segmentation image in HR. (c) HR segmentation image obtained by PSPNet. (d) SR segmentation image obtained by SrcNet. (e) SR segmentation image obtained by DSRL. (f) SR segmentation image obtained by our CSBSR.}
        }
        \label{fig:crack3}
\end{figure*}

Fig.~\ref{fig:crack3} shows sample test images where no crack pixels are observed.
While there are no crack pixels in these images, observed masonry joints tend to be false-positives.
For real applications using automatic image inspection, it is important to successfully suppress such false-positives for avoiding false alarms because most images have no crack pixels in real buildings.
In Fig.~\ref{fig:crack3}, it can be seen that (d) SrcNet and (e) DSRL detect false-positives around the masonry joints, while (f) CSBSR successfully neglects all of these masonry joint pixels.

\begin{table*}[ht!]
  \centering
  \caption{
  Performance change depending on $\beta$.
  $\beta$ is fixed during Step 3 in the training strategy, except for ``Increasing'' shown in the bottom line in which $\beta$ is increased from 0 to 1 in proportion to iterations.
  }
  \label{table:task_weight_change}
    \small
    \begin{tabular}{lccccccc}
    \toprule 
    {} & {} & \multicolumn{4}{c}{Segmentation metrics} & \multicolumn{2}{c}{SR metrics} \\
    {Model}  & $\beta$ & {IoU$_{max}\uparrow$} & {AIU$\uparrow$} & {HD95$_{min}\downarrow$} & {AHD95$\downarrow$} & {PSNR$\uparrow$} & {SSIM$\uparrow$} \\ 
    \cline{1-8}
    \midrule
    \multirow{8}{*}{\shortstack{\textbf{CSBSR} \\w/ PSPNet}} & w/o joint learning & 0.548 & 0.524 & 28.45 & 31.62 & 28.62 & \textcolor{red}{\textbf{0.706}} \\
    & 0.1 & 0.563 & 0.541 & 19.16 & 21.96 & 28.73 & 0.705 \\
     & 0.3 & \textcolor{red}{\textbf{0.573}} & \textcolor{red}{\textbf{0.552}} & 20.92 & 22.52 & \textcolor{red}{\textbf{28.75}} & 0.703 \\
     & 0.5 & 0.572 & 0.550 & 18.80 & 21.18 & 28.69 & 0.701 \\
    & 0.7 & 0.551 & 0.528 & 23.31 & 28.66 & 28.07 & 0.687 \\
     & 0.9 & 0.554 & 0.533 & 26.03 & 27.29 & 27.72 & 0.669 \\
     & 1.0 & 0.565 & 0.544 & 19.27 & 22.32 & 22.78 & 0.472 \\
     & Increasing & 0.568 & 0.549 & \textcolor{red}{\textbf{16.24}} & \textcolor{red}{\textbf{19.02}} & 27.12 & 0.662\\
    \midrule
    \multirow{8}{*}{\shortstack{\textbf{CSSR} \\ w/ PSPNet}} & w/o joint learning & 0.531 & 0.512 & 36.01 & 38.33 & 27.85 & \textcolor{red}{\textbf{0.667}} \\ 
    &0.1 & 0.547 & 0.529 & 24.45 & 28.27 & 28.42 & 0.653 \\
     & 0.3 & 0.475 & 0.446 & 53.75 & 55.96 & \textcolor{red}{\textbf{28.47}} & 0.663 \\
     & 0.5 & 0.546 & 0.523 & 22.12 & 24.61 & 28.39 & 0.657 \\
     & 0.7 & \textcolor{red}{\textbf{0.557}} & \textcolor{red}{\textbf{0.539}} & 21.20 & 24.74 & 28.35 & 0.656 \\
     & 0.9 & 0.552 & 0.534 & \textcolor{red}{\textbf{20.88}} & \textcolor{red}{\textbf{22.48}} & 28.01 & 0.653 \\
     & 1.0 & 0.539 & 0.515 & 21.82 & 26.04 & 20.29 & 0.436 \\
     &Increasing & 0.544 & 0.512 & 28.28 & 35.30 & 27.02 & 0.635 \\ 
    \bottomrule
    \end{tabular}
\end{table*}

\subsubsection{Element-wise Impact Verification}\label{subsubsec:element-verification}
\hfill\\
\noindent {\bf Effects of $\beta$:}
Table~\ref{table:task_weight_change} shows the evaluation results obtained in accordance with changes in $\beta$.
In all metrics of both SR and segmentation tasks, CSBSR outperforms CSSR.
Furthermore, in both CSSR and CSBSR, our proposed joint learning acquires better results in all segmentation metrics.

More specifically, in terms of the segmentation results, 
IoU$_{max}$ and AIU are not so changed depending on $\beta$.
On the other hand, the best HD95$_{min}$ and AHD95 scores are better in the training strategy with increasing $\beta$ (i.e., ``Increasing'' in the table) and have a larger margin from the scores obtained with any fixed $\beta$.
Intuitively speaking, the segmentation score should be best with $\beta=1$ so that the segmentation loss (i.e., $\mathcal{C}$ in Eq.~\ref{eq:joint_loss}) is fully weighted.
We interpret that the segmentation scores are not best with $\beta=1$ because it is difficult to fully optimize the whole network directly from the pre-trained SR and segmentation networks.
That is why the training strategy with increasing $\beta$ is better than $\beta=1$.

In terms of the SR image quality, 
While the best SSIM is acquired without joint learning, the best PSNR is with $\beta=0.3$.
Since the SR network is trained without joint learning just to improve SR, it is expected that the best SR results are obtained without joint learning.
This expectation is betrayed probably because of the feature extractor augmentation through the training of the segmentation task.
The features can be marginally augmented also for SR as in multi-task learning if $\beta$ is smaller, while the features are optimized for the segmentation task if $\beta$ is larger.


\noindent {\bf Effects of Segmentation losses:}
To verify the effectiveness of our BC and GBC losses, CSBSR is trained with other losses
for class-imbalance segmentation 
(i.e., WCE~\cite{bib:WCE_crack}, Dice~\cite{bib:Dice}, Combo~\cite{bib:taghanaki2019combo}, and GDice~\cite{DBLP:conf/miccai/SudreLVOC17}).
As shown in Table~\ref{table:loss_sota}, BC loss gets the best scores in four metrics (i.e., IoU, AIU, AHD95, and PSNR) and the second-best in HD95.
While it is the third place in SSIM, the gap from the best is tiny (0.705 vs 0.703).

\begin{figure*}[t]
\hspace{2mm}
\begin{minipage}[b]{0.47\linewidth}
  \centering
    ~\hspace{-6mm}
    \includegraphics[width=\linewidth]{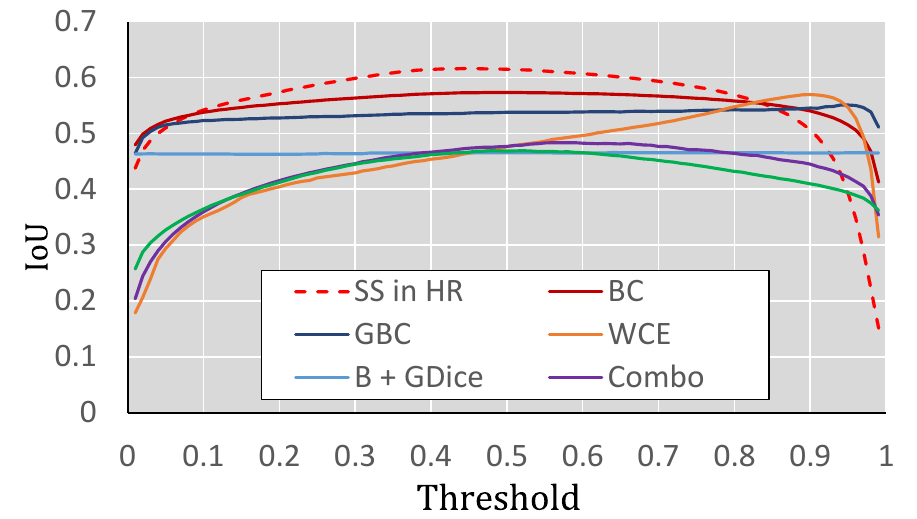}
    \vspace{-1em}
  \caption{
  Curves of IoU scores varying with a change in the threshold for segmentation image binarization. ``SS in HR'' means HR segmentation using BC loss. ``BC'' means SR segmentation using BC loss. ``GBC'' means SR segmentation using GBC loss. ``WCE'' means SR segmentation using WCE loss. ``B+GDice'' means SR segmentation using Boundary loss and GDice loss. ``Combo'' means SR segmentation using Combo loss. The segmentation model used is PSPNet for all conditions.}
  \label{fig:loss_curve_iou}
  \end{minipage}
  \hfill
  \begin{minipage}[b]{0.47\linewidth}

  \centering
    ~\hspace{-5mm}
    \includegraphics[width=\linewidth]{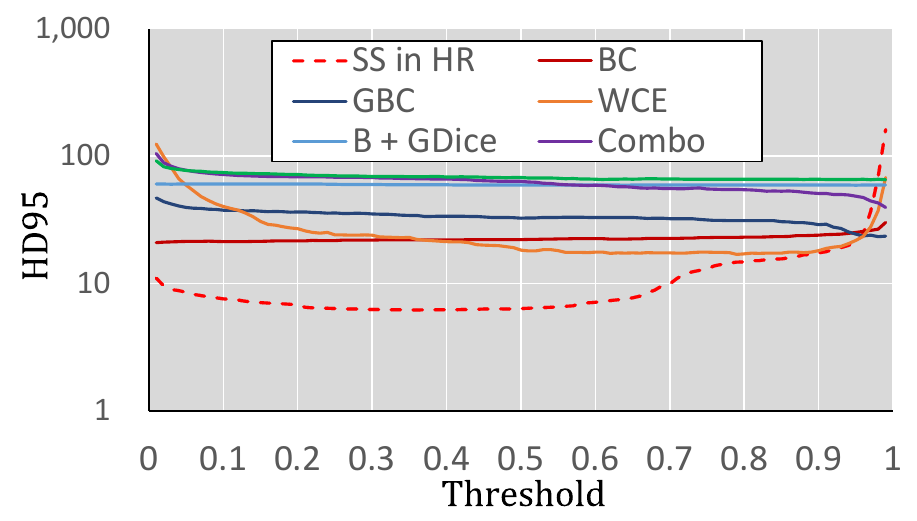}
    \vspace{-1em}
  \caption{
  Curves of HD95 scores varying with a change in the threshold for segmentation image binarization. \Erase{``SS in HR'' means HR segmentation using BC loss. ``BC'' means SR segmentation using BC loss. ``GBC'' means SR segmentation using GBC loss. ``WCE'' means SR segmentation using WCE loss. ``B+GDice'' means SR segmentation using Boundary loss and GDice loss. ``Combo'' means SR segmentation using Combo loss. The segmentation model used is PSPNet for all conditions.}
  }
  \label{fig:loss_curve_hd95}
\end{minipage}
\hspace{2mm}
\end{figure*}

Fig.~\ref{fig:loss_curve_iou} and Fig.~\ref{fig:loss_curve_hd95} shows IoU and HD95 scores varying with a change in a threshold for binarizing the segmentation image. 
As shown in Table~\ref{table:loss_sota}, GBC is inferior to BC.
However, GBC gets higher scores in a large range of thresholds in both IoU and HD95.
This property might be given by GDice, included in GBC, which works robustly to class imbalance.
On the other hand, while WCE gets better results in a few metrics in Table~\ref{table:loss_sota}, its performance drop depending on the threshold is significant.
This performance drop makes it difficult to apply WCE loss to a variety of scenarios.
As with GBC, the curves of BC are also not so decreased.

Based on the aforementioned observations, we conclude that our BC and GBC loesses are superior to other SoTA losses in terms of the max performance (as shown in Table~\ref{table:loss_sota}) and stability (as shown in Fig.~\ref{fig:loss_curve_iou} and Fig.~\ref{fig:loss_curve_hd95}.

\begin{table*}[t]
  \centering
  \caption{
  Comparison with other losses for class-imbalance segmentation.
  The best and second best scores are colored by \textcolor{red}{\textbf{red}} and \textcolor{blue}{\textbf{blue}}, respectively.
  }
  \label{table:loss_sota}
  \small
    \begin{tabular}{lcccccc} 
    \toprule 
    {} & \multicolumn{4}{c}{Segmentation metrics} & \multicolumn{2}{c}{SR metrics} \\
    {Model} & {IoU$_{max}\uparrow$} & {AIU$\uparrow$} & {HD95$_{min}\downarrow$} & {AHD95$\downarrow$} & {PSNR$\uparrow$} & {SSIM$\uparrow$}\\ \cline{1-7}
    \midrule
    \textbf{BC loss (Ours)} & \textcolor{red}{\textbf{0.573}} & \textcolor{red}{\textbf{0.552}} &  \textcolor{blue}{\textbf{20.92}} & \textcolor{red}{\textbf{22.52}} & \textcolor{red}{\textbf{28.75}} & 0.703 \\
    \textbf{GBC loss (Ours)} & 0.551 &  \textcolor{blue}{\textbf{0.534}} & 23.34 & 33.46 &  \textcolor{blue}{\textbf{28.70}} & \textcolor{red}{\textbf{0.705}} \\
    WCE~\cite{bib:WCE_crack} & \textcolor{blue}{\textbf{0.569}} & 0.459 & \textcolor{red}{\textbf{16.91}} &  \textcolor{blue}{\textbf{26.29}} & 28.60 & 0.704 \\
    Dice~\cite{bib:Dice} & 0.466 & 0.465 & 59.21 & 59.65 & 28.66 &  \textcolor{blue}{\textbf{0.704}} \\
    Combo~\cite{bib:taghanaki2019combo} & 0.483 & 0.436 & 39.48 & 62.27 & 28.51 & 0.697 \\
    Boundary~\cite{DBLP:journals/mia/KervadecBDGDA21} + GDice~\cite{DBLP:conf/miccai/SudreLVOC17} & 0.469 & 0.425 & 65.13 & 68.90 & 28.31 & 0.692 \\
    \bottomrule
    \end{tabular}
\end{table*}

\noindent {\bf Effects of Segmentation-aware SR-loss Weights:}
The effects of additional weights given to $\mathcal{L}_{S}$, which are proposed in Sec.~\ref{subsection:method_intermediate}, are evaluated in Table~\ref{table:exp_weights}.
Since $w^{C}$ and $w^{F}$ have hyper parameters (i.e., $m^{C}$ and $m^{F}$, respectively), the best results among $\left\{ m^{C}, m^{F} \right\} = \left\{ 2^{-3}, 2^{-2}, 2^{-1}, 2^{0}, 2^{1}, 2^{2}, 2^{3} \right\}$ are shown in Table~\ref{table:exp_weights}.
We can see the following observations:
\begin{itemize}
    \item All weights given to $\mathcal{L}_{S}$ improve HD95.
    \item Conversely, all weights given to $\mathcal{L}_{S}$ decrease IoU and AIU, while the performance drops are not so significant. In particular, IoU and AIU provided by $w^{F}$ given to $\mathcal{L}_{S}$ are almost equal to those of the baseline CSBSR (i.e., 0.573 vs 0.573 in IoU and 0.551 vs 0.552 in AIU).
    \item While $w^{F}$ weights the segmentation loss ($\mathcal{L}_{C}$), the results are inferior to the baseline in most metrics, as shown in the bottom row of Table~\ref{table:exp_weights}.
\end{itemize}

\begin{table*}[t]
  \centering
  \caption{Ablation study of weights given to $\mathcal{L}_{S}$, namely $\mathcal{L}_{C}$, $w^{C}$, and $w^{F}$.
  Scores better than the baseline (i.e., CSBSR without any weight) are underlined.
  }
  \label{table:exp_weights}
  \small
    \begin{tabular}{lcccccc} 
    \toprule 
    {} & \multicolumn{4}{c}{Segmentation metrics} & \multicolumn{2}{c}{SR metrics} \\
    {Model} & {IoU$_{max}\uparrow$} & {AIU$\uparrow$} & {HD95$_{min}\downarrow$} & {AHD95$\downarrow$} & {PSNR$\uparrow$} & {SSIM$\uparrow$}\\ 
    \cline{1-7}
    \midrule
    CSBSR & 0.573 & 0.552 & 20.92 & 22.52 & 28.75 & 0.703\\
    \midrule
w/ $L_{C}$ & 0.558 & 0.535 & \underline{19.72} & 22.90 & 27.32 & 0.649 \\
w/ $w^{C}$ ($m^{C}=8.0$) & 0.553 & 0.531 & \underline{19.21} & 26.02 & 28.70 & \underline{0.703}\\
w/ $w^{F}$ ($m^{F}=1.0$) & 0.573 & 0.551 & \textcolor{red}{\textbf{\underline{18.73}}} & \textcolor{red}{\textbf{\underline{21.70}}} & 28.73 & 0.702\\
w/ $w^{F}$ ($m^{F}=0.5$) for $\mathcal{L}_{C}$ & 0.556 & 0.531 & 22.26 & 25.94 & 28.70 & \textcolor{red}{\textbf{\underline{0.706}}}\\
    \bottomrule
    \end{tabular}
    \vspace{-1em}
\end{table*}

\begin{figure*}[t]
\begin{center}
\includegraphics[width=\textwidth]{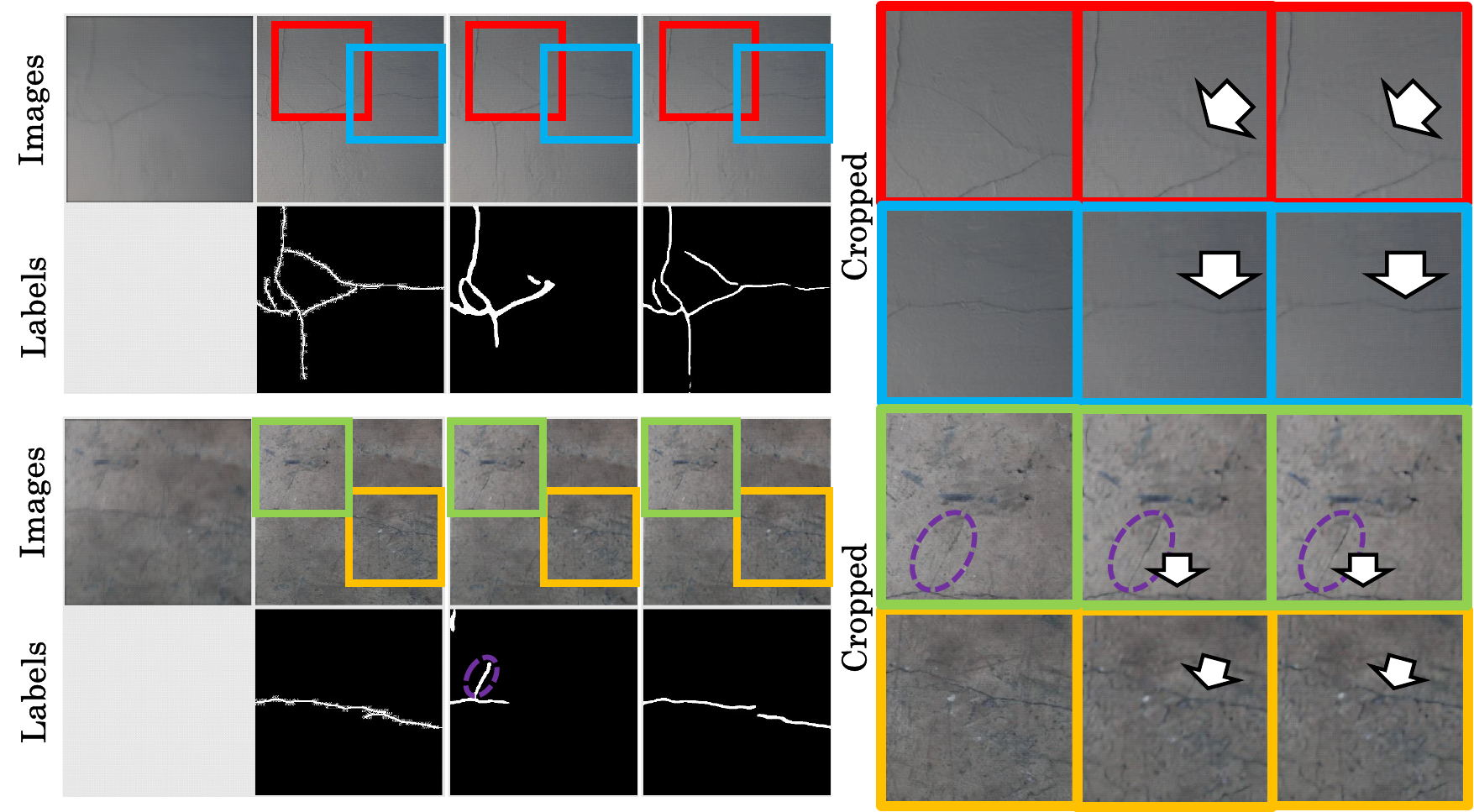}\\
~\hspace*{6mm}
(a) Input LR~\hspace*{4mm}
(b) HR GT~\hspace*{6mm}
(c) w/o $w^{F}$~\hspace*{6mm}
(d) w/ $w^{F}$~\hspace*{10mm}
(b') HR GT~\hspace*{5mm}
(c') w/o $w^{F}$~\hspace*{6mm}
(d') w/o $w^{F}$
\end{center}
\vspace{-1em}
\caption{
Visual comparison between CSBSR with and without the FO weight $w^{F}$.
[Left part] In the upper row of each example: (a) Input LR image (enlarged by Bicubic interpolation for visualization). (b) GT HR image. (c) SR image obtained by CSBSR without $w^{F}$. (d) SR image obtained by CSBSR.
In the lower row of each example: (a) No image. (b) GT segmentation image in HR. (c) SR segmentation image obtained by CSBSR without $w^{F}$. (d) SR segmentation image obtained by CSBSR.
[Right part] Rectangle regions are cropped from the SR images shown in the left part, and their zoom-in images are shown.
The boundary color of each cropped image shows the correspondence between the cropped images in the left and right parts.
Differences between (c') and (d') are pointed by white arrows.
}
\label{fig:ow_enhance}
\end{figure*}

In addition to the quantitative comparison shown in Table~\ref{table:exp_weights}, Fig.~\ref{fig:ow_enhance} visually shows the effect of the FO weight.
All images are the results obtained with $w^{F}=1.0$.
In the left part of Fig.~\ref{fig:ow_enhance}, we can see that $w^{F}$ allows CSBSR to detect thin crack pixels in segmentation images.
In order to see the results of SR image enhancement by $w^{F}$, the zoom-in images of several regions in the SR images are shown in the right part of Fig.~\ref{fig:ow_enhance}. In (c) images obtained without $w^{F}$, detected crack pixels are broken.
In (d) images obtained with $w^{F}$, on the other hand, cracks are more continuously detected, though it is difficult to visually see any significant difference between zoom-in SR images shown in (c') and (d').
In an opposite way, background textures enclosed by the purple dashed ellipse are falsely detected in CSBSR without $w^{F}$, as shown in (c) of the lower example.
However, these background pixels reconstructed by CSBSR without and with $w^{F}$ (enclosed by the purple dashed ellipses in (c') and (d')) are also almost the same as each other.
These results demonstrate the effectiveness of $w^{F}$ for discriminating between remarkably-similar crack and background pixels in the segmentation network of CSBSR.

\begin{figure*}[t]
\begin{center}
\includegraphics[width=\textwidth]{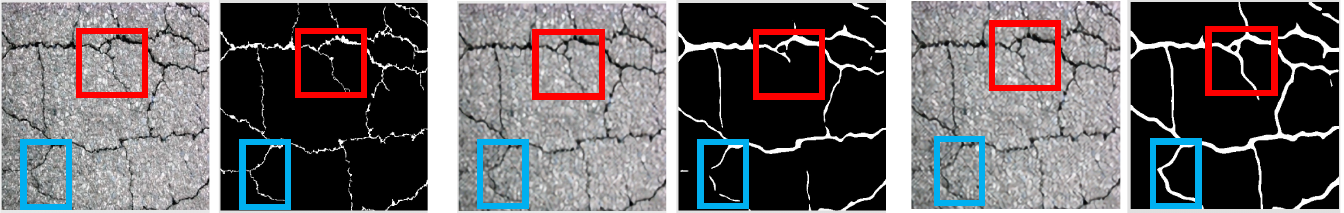}
~\hspace*{8mm}
(a) GT ~\hspace*{31mm}
(b) Results without blur skip ~\hspace*{17mm}
(c) Results with blur skip
\end{center}
\caption{Effectiveness of our proposed blur skip. (a) GT of SR and segmentation. (b) Results without blur skip of SR and segmentation. (c) Results with blur skip of SR and segmentation.
The left and right images show the HR/SR image and the segmentation image, respectively.
}
\label{fig:blur_skip_effect}
\end{figure*}

\noindent {\bf Effects of Blur Skip:}
The effects of the proposed blur skip process are shown in Table~\ref{table:exp_kernel_skip}.
Since the quality of the estimated kernel is high enough (e.g., above 50 dB in PSNR), our blur skip should have the potential to support the segmentation task.
While the single usage of the blur skip cannot work well for all metrics, the blur skip used with $w^{F}$ improves HD95 and AHD95.
The typical examples are shown in Fig.~\ref{fig:blur_skip_effect}.
While the results without the blur skip are much inferior to their GTs, the blur skip can improve the performance, as shown in the rightmost image in Fig.~\ref{fig:blur_skip_effect}.

\begin{table*}
  \centering
  \caption{
  Ablation study of our blur skip process.
  Scores better than the baseline (i.e., CSBSR without any weight) are underlined.
  }
  \label{table:exp_kernel_skip}
  \small
    \begin{tabular}{lccccccc} 
    \toprule 
    {} & \multicolumn{4}{c}{Segmentation metrics} & \multicolumn{3}{c}{SR metrics} \\
    {Model} & {IoU$_{max}\uparrow$} & {AIU$\uparrow$} & {HD95$_{min}\downarrow$} & {AHD95$\downarrow$} & {PSNR$\uparrow$} & {SSIM$\uparrow$} & {Kernel PSNR$\uparrow$}\\ 
    \cline{1-8}
    \midrule
    CSBSR & 0.573 & 0.552 & 20.92 & 22.52 & 28.75 & 0.703 & 50.65\\
    \midrule
CSBSR w/ KS & 0.544 & 0.523 & 28.86 & 32.02 & 28.52 & 0.696 &  \underline{50.82} \\
CSBSR w/ KS and $m^{F}=1.0$ & 0.550 & 0.528 & \textcolor{red}{\textbf{\underline{18.06}}} & \textcolor{red}{\textbf{\underline{19.10}}} & 28.65 & 0.702 &  \textcolor{red}{\textbf{\underline{50.91}}}\\
    \bottomrule
    \end{tabular}
\end{table*}

\subsection{Crack Images with Real Degradations}
\label{subsection:exp_real}

For experiments with real images, we captured 809 wall images ($1280\times720$ pixels) with a flying drone (DJI MAVIC MINI).
This dataset includes out-of-focus images as well as motion-blurred images.
By using all the images in this dataset as test images, we visually verify the effectiveness of CSBSR for realistically-blurred images.
Since it is essentially difficult to annotate severely-blurred cracks correctly, only qualitative comparison is done with this dataset.

\begin{figure*}[t]
  \centering
    \includegraphics[width=\textwidth]{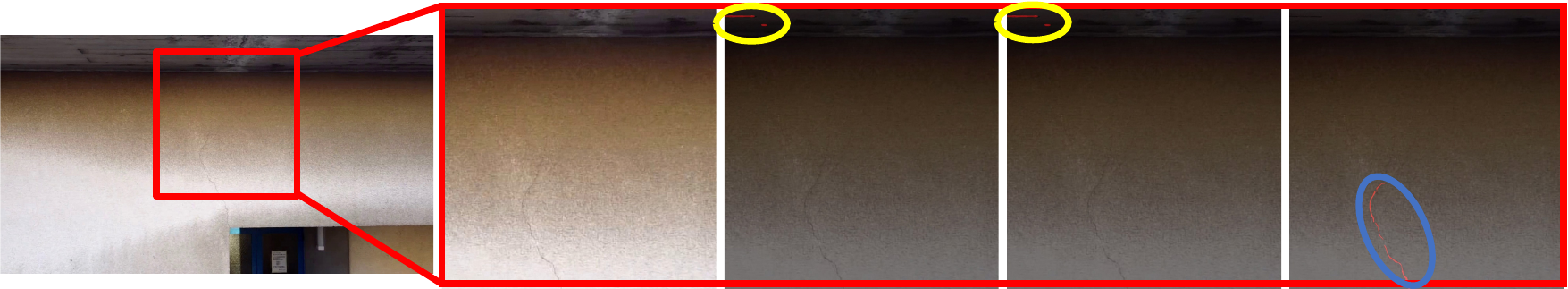}
    \includegraphics[width=\textwidth]{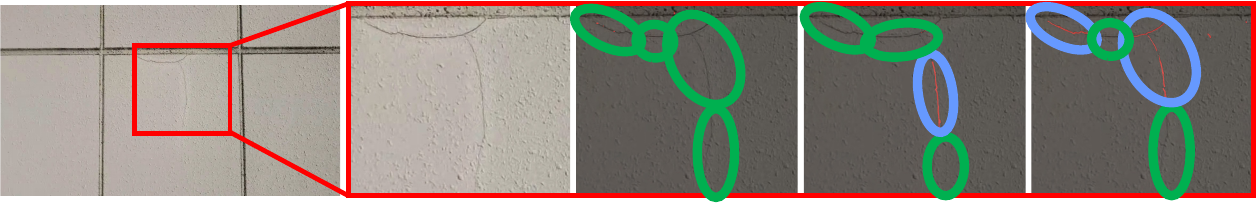}
    \includegraphics[width=\textwidth]{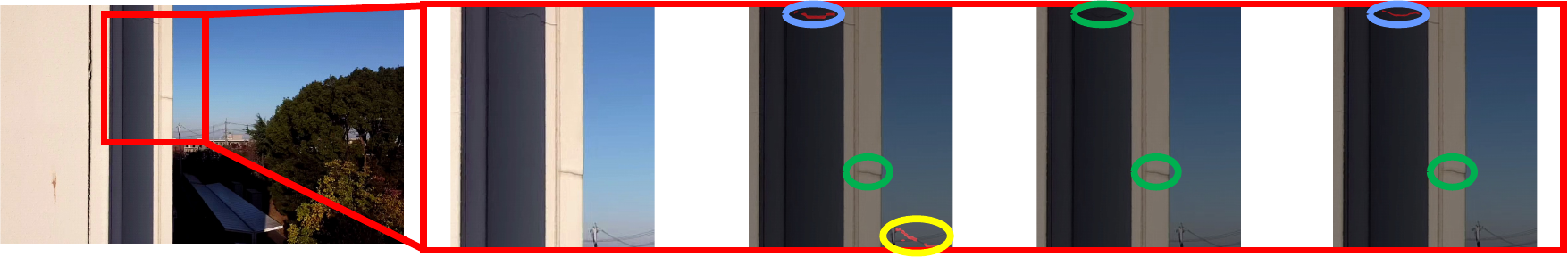}
    ~\hspace*{1mm}
    (a) Input image \hspace*{10mm}
    (b) Cropped input image \hspace*{4mm}
    (c) DSRL~\cite{DBLP:conf/cvpr/WangLZTS20} \hspace*{8mm}
    (d) SrcNet~\cite{bib:SrcNet} \hspace*{12mm}
    (e) Ours
  \caption{Visual results of comparative experiments on real blurred crack images. (a) Input LR image. (b) Cropped and Enlarged input image. In (c), (d), and (e), detected crack pixel are colored by red. True-positive, false-negative, and false-positive cracks are enclosed by blue, green, and yellow ellipses, respectively.}
  \label{fig:real}
  \vspace{-1em}
\end{figure*}

In the first row of Fig.~\ref{fig:real}, cracks are very thin.
DSRL and SrcNet cannot detect any crack pixels.
In addition, false-positive cracks  (enclosed by yellow ellipses) are detected.
CSBSR, on the other hand, can detect most crack pixels, as depicted by superimposed red pixels.

The second row of Fig.~\ref{fig:real} shows the segmentation results detected on the image of complex cracks observed on a building wall.
While DSRL detects no crack pixels, SrcNet and CSBSR successfully detect several crack pixels.
CSBSR can detect more true-positive crack pixels, in particular, along a crack located in the upper part of the image (enclosed by blue ellipses).
However, there are also many false-negative crack pixels (enclosed by green ellipses) even in the segmentation image of CSBSR.

In the input image shown in the third row of Fig.~\ref{fig:real}, there are thin electrical wires as well as thin cracks (enclosed by blue and green ellipses).
A crack segmentation method is required to detect only real cracks without being disturbed by the wires.
DSRL detects several wire pixels (enclosed by the yellow ellipse) and crack pixels, while SrcNet detects nothing.
While CSBSR detects only crack pixels, even CSBSR fails to detect blurry cracks observed in the lower part of the image (enclosed by green ellipses).

As mentioned above, while our CSBSR outperforms SoTA segmentation methods using SR, it also fails to detect severely-degraded cracks.
Improving crack segmentation in such severely-degraded images is important for future work.

\section{Concluding Remarks}
\label{section:conclusion}

This paper proposes an end-to-end joint learning network consisting of blind SR and segmentation networks.
Blind SR allows us to apply the proposed method to realistically-blurred images.
The information exchange between the SR and segmentation networks (i.e., segmentation-aware SR-loss weights and blur skip for blur-reflected task learning) enables further improvement.
For better segmentation in class-imbalance fine crack images, BC loss is proposed.

Future work includes quantitative evaluation on real-image datasets in which GT segmentation pixels are manually given.
It is also interesting to apply CSBSR to other segmentation tasks such as medical imaging.
An essential difficulty in SR is that SR is an ill-posed problem in which a larger number of pixels are reconstructed from a smaller number of pixels.
In order to relieve this difficulty, multiple LR images are used as a
set of input images in video
SR~\cite{DBLP:conf/cvpr/HarisSU20}.
Our proposed method can also be extended to the one with time-series images.
Hyperparameter optimization may be also useful for exploring better parameters, while several parameters such as $\beta$ and $\gamma$ are empirically determined.

\section*{Acknowledgments}

This work was partly supported by JSPS KAKENHI Grant Numbers 19K12129 and 22H03618.

\bibliographystyle{IEEEtran}

\bibliography{ref-csbsr}

\begin{thebibliography}{10}
\providecommand{\url}[1]{#1}
\csname url@samestyle\endcsname
\providecommand{\newblock}{\relax}
\providecommand{\bibinfo}[2]{#2}
\providecommand{\BIBentrySTDinterwordspacing}{\spaceskip=0pt\relax}
\providecommand{\BIBentryALTinterwordstretchfactor}{4}
\providecommand{\BIBentryALTinterwordspacing}{\spaceskip=\fontdimen2\font plus
\BIBentryALTinterwordstretchfactor\fontdimen3\font minus \fontdimen4\font\relax}
\providecommand{\BIBforeignlanguage}[2]{{%
\expandafter\ifx\csname l@#1\endcsname\relax
\typeout{** WARNING: IEEEtran.bst: No hyphenation pattern has been}%
\typeout{** loaded for the language `#1'. Using the pattern for}%
\typeout{** the default language instead.}%
\else
\language=\csname l@#1\endcsname
\fi
#2}}
\providecommand{\BIBdecl}{\relax}
\BIBdecl

\bibitem{bib:crack_survey1}
F.~Elghaish, S.~Talebi, E.~Abdellatef, S.~T. Matarneh, M.~R. Hosseini, S.~Wu, M.~Mayouf, A.~Hajirasouli \emph{et~al.}, ``Developing a new deep learning cnn model to detect and classify highway cracks,'' \emph{Journal of Engineering, Design and Technology}, 2021.

\bibitem{DBLP:journals/pami/ShelhamerLD17}
E.~Shelhamer, J.~Long, and T.~Darrell, ``Fully convolutional networks for semantic segmentation,'' \emph{{IEEE} Trans. Pattern Anal. Mach. Intell.}, vol.~39, no.~4, pp. 640--651, 2017.

\bibitem{DBLP:journals/pami/HeGDG20}
K.~He, G.~Gkioxari, P.~Doll{\'{a}}r, and R.~B. Girshick, ``Mask {R-CNN},'' \emph{{IEEE} Trans. Pattern Anal. Mach. Intell.}, vol.~42, no.~2, pp. 386--397, 2020.

\bibitem{DBLP:conf/cvpr/KirillovHGRD19}
A.~Kirillov, K.~He, R.~B. Girshick, C.~Rother, and P.~Doll{\'{a}}r, ``Panoptic segmentation,'' in \emph{CVPR}, 2019.

\bibitem{DBLP:conf/mva/StentGSSC13}
S.~Stent, R.~Gherardi, B.~Stenger, K.~Soga, and R.~Cipolla, ``An image-based system for change detection on tunnel linings,'' in \emph{MVA}, 2013.

\bibitem{DBLP:journals/tits/FeiWZCLLYL20}
Y.~Fei, K.~C.~P. Wang, A.~Zhang, C.~Chen, J.~Q. Li, Y.~Liu, G.~Yang, and B.~Li, ``Pixel-level cracking detection on 3d asphalt pavement images through deep-learning- based cracknet-v,'' \emph{{IEEE} Trans. Intell. Transp. Syst.}, vol.~21, no.~1, pp. 273--284, 2020.

\bibitem{CSSR2021}
Y.~Kondo and N.~Ukita, ``Crack segmentation for low-resolution images using joint learning with super-resolution,'' in \emph{MVA}, 2021.

\bibitem{bib:SrcNet}
H.~Bae, K.~Jang, and Y.-K. An, ``Deep super resolution crack network (srcnet) for improving computer vision--based automated crack detectability in in situ bridges,'' \emph{Structural Health Monitoring}, vol.~20, no.~4, pp. 1428--1442, 2021.

\bibitem{DBLP:conf/cvpr/WangLZTS20}
L.~Wang, D.~Li, Y.~Zhu, L.~Tian, and Y.~Shan, ``Dual super-resolution learning for semantic segmentation,'' in \emph{CVPR}, 2020, \url{https://github.com/Dootmaan/DSRL}.

\bibitem{DBLP:journals/corr/abs-2001-05566}
S.~Minaee, Y.~Boykov, F.~Porikli, A.~Plaza, N.~Kehtarnavaz, and D.~Terzopoulos, ``Image segmentation using deep learning: {A} survey,'' \emph{arXiv}, 2020.

\bibitem{bib:WCE_crack}
D.~Dais, İhsan Engin~Bal, E.~Smyrou, and V.~Sarhosis, ``Automatic crack classification and segmentation on masonry surfaces using convolutional neural networks and transfer learning,'' \emph{Automation in Construction}, vol. 125, p. 103606, 2021.

\bibitem{DBLP:conf/iccv/LinGGHD17}
T.~Lin, P.~Goyal, R.~B. Girshick, K.~He, and P.~Doll{\'{a}}r, ``Focal loss for dense object detection,'' in \emph{ICCV}, 2017.

\bibitem{DBLP:journals/tmi/LiKG21}
Z.~Li, K.~Kamnitsas, and B.~Glocker, ``Analyzing overfitting under class imbalance in neural networks for image segmentation,'' \emph{{IEEE} Trans. Medical Imaging}, vol.~40, no.~3, pp. 1065--1077, 2021.

\bibitem{DBLP:journals/ijon/HossainBP21}
M.~S. Hossain, J.~M. Betts, and A.~P. Paplinski, ``Dual focal loss to address class imbalance in semantic segmentation,'' \emph{Neurocomputing}, vol. 462, pp. 69--87, 2021.

\bibitem{DBLP:conf/cvpr/BuloNK17}
S.~R. Bul{\`{o}}, G.~Neuhold, and P.~Kontschieder, ``Loss max-pooling for semantic image segmentation,'' in \emph{CVPR}, 2017.

\bibitem{DBLP:journals/cvm/GongZZYX22}
L.~Gong, Y.~Zhang, Y.~Zhang, Y.~Yang, and W.~Xu, ``Erroneous pixel prediction for semantic image segmentation,'' \emph{Comput. Vis. Media}, vol.~8, no.~1, pp. 165--175, 2022.

\bibitem{bib:Dice}
F.~Milletari, N.~Navab, and S.~Ahmadi, ``V-net: Fully convolutional neural networks for volumetric medical image segmentation,'' in \emph{3DV}, 2016.

\bibitem{DBLP:conf/miccai/SudreLVOC17}
C.~H. Sudre, W.~Li, T.~Vercauteren, S.~Ourselin, and M.~J. Cardoso, ``Generalised dice overlap as a deep learning loss function for highly unbalanced segmentations,'' in \emph{MICCAI}, 2017.

\bibitem{bib:taghanaki2019combo}
S.~A. Taghanaki, Y.~Zheng, S.~K. Zhou, B.~Georgescu, P.~Sharma, D.~Xu, D.~Comaniciu, and G.~Hamarneh, ``Combo loss: Handling input and output imbalance in multi-organ segmentation,'' \emph{Comput Med Imaging Graph}, vol.~75, pp. 24--33, 2019.

\bibitem{DBLP:journals/tmi/KarimiS20}
D.~Karimi and S.~E. Salcudean, ``Reducing the hausdorff distance in medical image segmentation with convolutional neural networks,'' \emph{{IEEE} Trans. Medical Imaging}, vol.~39, no.~2, pp. 499--513, 2020.

\bibitem{DBLP:journals/mia/KervadecBDGDA21}
H.~Kervadec, J.~Bouchtiba, C.~Desrosiers, E.~Granger, J.~Dolz, and I.~B. Ayed, ``Boundary loss for highly unbalanced segmentation,'' \emph{Medical Image Anal.}, vol.~67, p. 101851, 2021.

\bibitem{bib:Zhang2020}
K.~Zhang, Y.~Zhang, and H.-D. Cheng, ``Crackgan: Pavement crack detection using partially accurate ground truths based on generative adversarial learning,'' \emph{TITS}, vol.~22, no.~2, pp. 1306--1319, 2020.

\bibitem{bib:rezaie2020}
A.~Rezaie, R.~Achanta, M.~Godio, and K.~Beyer, ``Comparison of crack segmentation using digital image correlation measurements and deep learning,'' \emph{Construction and Building Materials}, vol. 261, no.~20, p. 120474, 2020.

\bibitem{bib:chen2021}
H.~Chen, Y.~Su, and W.~He, ``Automatic crack segmentation using deep high-resolution representation learning,'' \emph{Applied Optics}, vol.~60, no.~21, pp. 6080--6090, 2021.

\bibitem{bib:liu2019deepcrack}
Y.~Liu, J.~Yao, X.~Lu, R.~Xie, and L.~Li, ``Deepcrack: A deep hierarchical feature learning architecture for crack segmentation,'' \emph{Neurocomputing}, vol. 338, pp. 139--153, 2019.

\bibitem{bib:liu2021crackformer}
H.~Liu, X.~Miao, C.~Mertz, C.~Xu, and H.~Kong, ``Crackformer: Transformer network for fine-grained crack detection,'' in \emph{ICCV}, 2021.

\bibitem{DBLP:journals/pami/ChenPKMY18}
L.~Chen, G.~Papandreou, I.~Kokkinos, K.~Murphy, and A.~L. Yuille, ``Deeplab: Semantic image segmentation with deep convolutional nets, atrous convolution, and fully connected crfs,'' \emph{{IEEE} Trans. Pattern Anal. Mach. Intell.}, vol.~40, no.~4, pp. 834--848, 2018.

\bibitem{bib:Ronneberger2015unet}
O.~Ronneberger, P.~Fischer, and T.~Brox, ``U-net: Convolutional networks for biomedical image segmentation,'' in \emph{MICCAI}, 2015.

\bibitem{bib:Yang2020}
F.~Yang, L.~Zhang, S.~Yu, D.~Prokhorov, X.~Mei, and H.~Ling, ``Feature pyramid and hierarchical boosting network for pavement crack detection,'' \emph{TITS}, vol.~21, no.~4, pp. 1525--1535, 2020.

\bibitem{bib:wang2021SR-survey}
Z.~Wang, J.~Chen, and S.~C.~H. Hoi, ``Deep learning for image super-resolution: {A} survey,'' \emph{TPAMI}, vol.~43, no.~10, pp. 3365--3387, 2021.

\bibitem{DBLP:conf/cvpr/ZhangGT20}
K.~Zhang, L.~V. Gool, and R.~Timofte, ``Deep unfolding network for image super-resolution,'' in \emph{CVPR}, 2020.

\bibitem{DBLP:conf/cvpr/LeePLMKLKHK20}
J.~Lee, J.~Park, K.~Lee, J.~Min, G.~Kim, B.~Lee, B.~Ku, D.~K. Han, and H.~Ko, ``{FBRNN:} feedback recurrent neural network for extreme image super-resolution,'' in \emph{CVPR Workshop}, 2020.

\bibitem{DBLP:conf/cvpr/LuLTLJ21}
L.~Lu, W.~Li, X.~Tao, J.~Lu, and J.~Jia, ``{MASA-SR:} matching acceleration and spatial adaptation for reference-based image super-resolution,'' in \emph{CVPR}, 2021.

\bibitem{bib:MZSR}
J.~W. Soh, S.~Cho, and N.~I. Cho, ``Meta-transfer learning for zero-shot super-resolution,'' in \emph{CVPR}, 2020.

\bibitem{bib:BSRGAN2021}
K.~Zhang, J.~L. L.~V. Gool, and R.~Timofte, ``Designing a practical degradation model for deep blind image super-resolution,'' in \emph{ICCV}, 2021.

\bibitem{bib:hussein2020correction}
S.~A. Hussein, T.~Tirer, and R.~Giryes, ``Correction filter for single image super-resolution: Robustifying off-the-shelf deep super-resolvers,'' in \emph{CVPR}, 2020.

\bibitem{DBLP:conf/cvpr/GuLZD19}
J.~Gu, H.~Lu, W.~Zuo, and C.~Dong, ``Blind super-resolution with iterative kernel correction,'' in \emph{CVPR}, 2019.

\bibitem{DBLP:conf/cvpr/WangWDX0AG21}
L.~Wang, Y.~Wang, X.~Dong, Q.~Xu, J.~Yang, W.~An, and Y.~Guo, ``Unsupervised degradation representation learning for blind super-resolution,'' in \emph{CVPR}, 2021.

\bibitem{DBLP:conf/cvpr/GuoCWCCDXT20}
Y.~Guo, J.~Chen, J.~Wang, Q.~Chen, J.~Cao, Z.~Deng, Y.~Xu, and M.~Tan, ``Closed-loop matters: Dual regression networks for single image super-resolution,'' in \emph{CVPR}, 2020.

\bibitem{DBLP:conf/cvpr/KimSK21}
S.~Y. Kim, H.~Sim, and M.~Kim, ``Koalanet: Blind super-resolution using kernel-oriented adaptive local adjustment,'' in \emph{CVPR}, 2021.

\bibitem{bib:kbpn}
T.~Yoshida, Y.~Kondo, T.~Maeda, K.~Akita, and N.~Ukita, ``Kernelized back-projection networks for blind super resolution,'' \emph{arXiv}, 2023.

\bibitem{DBLP:conf/cvpr/HaoLQYLH17}
Z.~Hao, Y.~Liu, H.~Qin, J.~Yan, X.~Li, and X.~Hu, ``Scale-aware face detection,'' in \emph{CVPR}, 2017.

\bibitem{DBLP:journals/access/GuoWSYCZSXXSS19}
Z.~Guo, G.~Wu, X.~Song, W.~Yuan, Q.~Chen, H.~Zhang, X.~Shi, M.~Xu, Y.~Xu, R.~Shibasaki, and X.~Shao, ``Super-resolution integrated building semantic segmentation for multi-source remote sensing imagery,'' \emph{{IEEE} Access}, vol.~7, pp. 99\,381--99\,397, 2019.

\bibitem{DBLP:journals/tits/ChoiCCS18}
D.~Choi, J.~H. Choi, J.~W. Choi, and B.~C. Song, ``Sharpness enhancement and super-resolution of around-view monitor images,'' \emph{{IEEE} Trans. Intell. Transp. Syst.}, vol.~19, no.~8, pp. 2650--2662, 2018.

\bibitem{DBLP:conf/iccv/SinghN0V19}
M.~Singh, S.~Nagpal, R.~Singh, and M.~Vatsa, ``Dual directed capsule network for very low resolution image recognition,'' in \emph{ICCV}, 2019.

\bibitem{DBLP:journals/tnn/ZhangBDXG21}
Y.~Zhang, Y.~Bai, M.~Ding, S.~Xu, and B.~Ghanem, ``Kgsnet: Key-point-guided super-resolution network for pedestrian detection in the wild,'' \emph{{IEEE} Trans. Neural Networks Learn. Syst.}, vol.~32, no.~5, pp. 2251--2265, 2021.

\bibitem{DBLP:conf/ipas2/AkitaHU20}
K.~Akita, M.~Haris, and N.~Ukita, ``Region-dependent scale proposals for super-resolution in object detection,'' in \emph{IPAS}, 2020.

\bibitem{bib:TDSR2021}
M.~Haris, G.~Shakhnarovich, and N.~Ukita, ``Task-driven super resolution: Object detection in low-resolution images,'' in \emph{ICONIP}, 2021.

\bibitem{bib:DBPN}
------, ``Deep back-projection networks for super-resolution,'' in \emph{CVPR}, 2018.

\bibitem{bib:zhao2017PSPNet}
H.~Zhao, J.~Shi, X.~Qi, X.~Wang, and J.~Jia, ``Pyramid scene parsing network,'' in \emph{CVPR}, 2017.

\bibitem{DBLP:conf/eccv/YuanCW20}
Y.~Yuan, X.~Chen, and J.~Wang, ``Object-contextual representations for semantic segmentation,'' in \emph{ECCV}, 2020, \url{https://github.com/openseg-group/openseg.pytorch/blob/master/MODEL_ZOO.md}.

\bibitem{DBLP:journals/mia/MaCNHLLYM21}
J.~Ma, J.~Chen, M.~Ng, R.~Huang, Y.~Li, C.~Li, X.~Yang, and A.~L. Martel, ``Loss odyssey in medical image segmentation,'' \emph{Medical Image Anal.}, vol.~71, p. 102035, 2021.

\bibitem{DBLP:conf/iccv/RyouJP19}
S.~Ryou, S.~Jeong, and P.~Perona, ``Anchor loss: Modulating loss scale based on prediction difficulty,'' in \emph{ICCV}, 2019.

\bibitem{bib:code_hd95}
DeepMind, ``Surface distance metrics,'' \url{https://github.com/deepmind/surface-distance}.

\bibitem{DBLP:conf/cvpr/WangYDL18}
X.~Wang, K.~Yu, C.~Dong, and C.~C. Loy, ``Recovering realistic texture in image super-resolution by deep spatial feature transform,'' in \emph{CVPR}, 2018.

\bibitem{bib:ntire2017}
E.~Agustsson and R.~Timofte, ``Ntire 2017 challenge on single image super-resolution: Dataset and study,'' in \emph{CVPRW}, 2017.

\bibitem{bib:Flickr2K}
R.~Timofte, E.~Agustsson, L.~Van~Gool, M.-H. Yang, and L.~Zhang, ``Ntire 2017 challenge on single image super-resolution: Methods and results,'' in \emph{CVPRW}, 2017.

\bibitem{bib:deng2009imagenet}
J.~Deng, W.~Dong, R.~Socher, L.-J. Li, K.~Li, and L.~Fei-Fei, ``Imagenet: A large-scale hierarchical image database,'' in \emph{CVPR}, 2009.

\bibitem{DBLP:journals/corr/SimonyanZ14a}
K.~Simonyan and A.~Zisserman, ``Very deep convolutional networks for large-scale image recognition,'' in \emph{ICLR}, 2015.

\bibitem{bib:torchvision}
``Torchvision.models,'' \url{https://pytorch.org/vision/stable/models.html}.

\bibitem{Adam}
D.~P. Kingma and J.~Ba, ``Adam: {A} method for stochastic optimization,'' in \emph{ICLR}, 2015.

\bibitem{bib:Khanhha}
Khanhha, ``Crack segmentation,'' 2020, \url{https://github.com/khanhha/crack_segmentation}.

\bibitem{bib:zhang2016road}
L.~Zhang, F.~Yang, Y.~D. Zhang, and Y.~J. Zhu, ``Road crack detection using deep convolutional neural network,'' in \emph{ICIP}, 2016.

\bibitem{bib:eisenbach2017how}
M.~Eisenbach, R.~Stricker, D.~Seichter, K.~Amende, K.~Debes, M.~Sesselmann, D.~Ebersbach, U.~Stoeckert, and H.-M. Gross, ``How to get pavement distress detection ready for deep learning? a systematic approach.'' in \emph{IJCNN}, 2017.

\bibitem{bib:shi2016automatic}
Y.~Shi, L.~Cui, Z.~Qi, F.~Meng, and Z.~Chen, ``Automatic road crack detection using random structured forests,'' \emph{TITS}, vol.~17, no.~12, pp. 3434--3445, 2016.

\bibitem{bib:amhaz2016automatic}
R.~Amhazand, S.~Chambon, J.~Idier, and V.~Baltazart, ``Automatic crack detection on two-dimensional pavement images: An algorithm based on minimal path selection,'' \emph{TITS}, vol.~17, no.~10, pp. 2718--2729, 2016.

\bibitem{bib:zou2012cracktree}
Q.~Zou, Y.~Cao, Q.~Li, Q.~Mao, and S.~Wang, ``Cracktree: Automatic crack detection from pavement images,'' \emph{Pattern Recognition Letters}, vol.~33, no.~3, pp. 227--238, 2012.

\bibitem{bib:CSSC}
L.~Yang, B.~Li, W.~Li, L.~Zhaoming, G.~Yang, and J.~Xiao, ``Deep concrete inspection using unmanned aerial vehicle towards cssc database,'' in \emph{IROS}, 2017.

\bibitem{bib:SSIM}
Z.~Wang, A.~C. Bovik, H.~R. Sheikh, and E.~P. Simoncelli, ``Image quality assessment: from error visibility to structural similarity,'' \emph{IEEE Transactions on image processing}, vol.~13, no.~4, pp. 600--612, 2004.

\bibitem{DBLP:conf/cvpr/HarisSU20}
M.~Haris, G.~Shakhnarovich, and N.~Ukita, ``Space-time-aware multi-resolution video enhancement,'' in \emph{CVPR}, 2020.

\end{thebibliography}

\section*{Biography Section}


\begin{IEEEbiography}[{\includegraphics[width=1in,height=1.25in,clip,keepaspectratio]{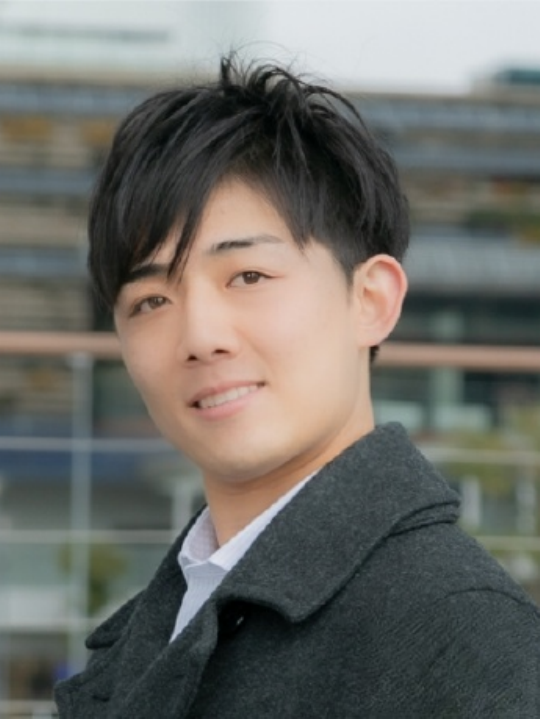}}]{Yuki Kondo} received the bachelor degree in engineering from Toyota Technological Institute in 2022. 
Currently, he is a researcher at Toyota Technological Institute. His research interests include low-level vision including image and video super-resolution and its application to tiny image analysis such as crack detection. His award includes the best practical paper award in MVA2021.
\end{IEEEbiography}

\begin{IEEEbiography}[{\includegraphics[width=1in,height=1.25in,clip,keepaspectratio]{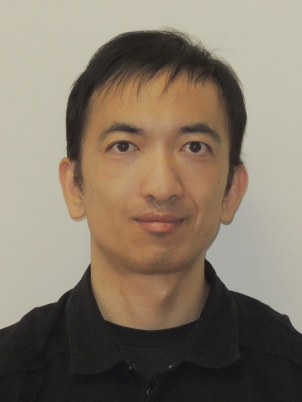}}]{Norimichi Ukita} received the B.E. and M.E. degrees in information engineering from Okayama University, Japan, in 1996 and 1998, respectively, and the Ph.D. degree in Informatics from Kyoto University, Japan, in 2001.
  From 2001 to 2016, he was an assistant professor (2001 to 2007) and an associate professor (2007-2016) with the graduate school of information science, Nara Institute of Science and Technology, Japan. In 2016, he became a professor at Toyota Technological Institute, Japan. He was a research scientist of Precursory Research for Embryonic Science and Technology, Japan Science and Technology Agency, during 2002--2006, and a visiting research scientist at Carnegie Mellon University during 2007--2009.
  Currently, he is also an adjunct professor at Toyota Technological Institute at Chicago.
  Prof. Ukita's awards include the excellent paper award of IEICE (1999), the winner award in NTIRE 2018 challenge on image super-resolution, 1st place in PIRM 2018 perceptual SR challenge, the best poster award in MVA2019, and the best practical paper award in MVA2021.
\end{IEEEbiography}

\vfill

\end{document}